\documentclass[10pt,journal,compsoc]{IEEEtran}

\ifCLASSOPTIONcompsoc
  \usepackage[nocompress]{cite}
\else
  \usepackage{cite}
\fi
\ifCLASSINFOpdf
\else
\fi
\usepackage{amsmath,amsfonts}
\usepackage{array}
\usepackage[caption=false,font=normalsize,labelfont=sf,textfont=sf]{subfig}
\usepackage{textcomp}
\usepackage{stfloats}
\usepackage{url}
\usepackage{verbatim}
\usepackage{graphicx}
\usepackage{cite}
\usepackage[misc]{ifsym}
\usepackage{algorithm}
\usepackage{algorithmic}
\usepackage{amsmath}
\usepackage{threeparttable}
\usepackage{graphicx}
\usepackage{cite}
\usepackage{color}
\usepackage{colortbl}
\usepackage{xcolor}
\usepackage{multirow}
\usepackage{booktabs}
\usepackage{amssymb}
\usepackage{pict2e}
\usepackage{threeparttable}
\usepackage{textcomp}
\usepackage{verbatim}
\usepackage{graphicx}
\usepackage{cite}
\usepackage{color}
\usepackage{colortbl}
\usepackage{xcolor}
\usepackage{multirow}
\usepackage{bbding}
\usepackage{caption}
\usepackage{footnote}
\usepackage{shadowtext}
\usepackage{booktabs}
\usepackage{subfig} 
\usepackage{subfloat}
\usepackage{multicol}
\usepackage{listings}
\usepackage{xcolor}
\usepackage{enumitem}
\usepackage{comment}
\usepackage{url}

\usepackage[T1]{fontenc}    % use 8-bit T1 fonts
\usepackage{url}            % simple URL typesetting
\usepackage{booktabs}       % professional-quality tables
\usepackage{amsfonts}       % blackboard math symbols
\usepackage{dsfont}
\usepackage{multirow}

\usepackage{nicefrac}       % compact symbols for 1/2, etc.
\usepackage{microtype}      % microtypography

\usepackage[colorlinks=false, pdfborder={0 0 0}]{hyperref}

\let\oldurl\url
\renewcommand{\url}[1]{\textcolor{green!50!black}{\oldurl{#1}}}

\newcommand{\RED}[1]{\textcolor{black}{#1}}

\begin{document}

\title{Foundation Model for Skeleton-Based Human Action Understanding}

\author{
Hongsong~Wang, Wanjiang~Weng, Junbo~Wang, Fang~Zhao, Guo-Sen~Xie, Xin~Geng, Senior Member, IEEE, and Liang Wang, Fellow, IEEE
\IEEEcompsocitemizethanks{
	\IEEEcompsocthanksitem H. Wang, W. Weng and X. Geng are with School of Computer Science and Engineering, Southeast University, Nanjing 211189, China, and also with Key Laboratory of New Generation Artificial Intelligence Technology and Its Interdisciplinary Applications (Southeast University), Ministry of Education, China (email: hongsongwang@seu.edu.cn; 220232322@seu.edu.cn; xgeng@seu.edu.cn).
	\IEEEcompsocthanksitem J. Wang is with School of Software, Northwestern Polytechnical University, Xi’an 710072, China (email: jbwang@nwpu.edu.cn).
	\IEEEcompsocthanksitem F. Zhao is with State Key Laboratory for Novel Software Technology and School of Intelligence Science and Technology, Nanjing University, Nanjing 210023, China (email: zhaofang0627@gmail.com). 
	\IEEEcompsocthanksitem G. Xie is with School of Computer Science and Engineering, Nanjing University of Science and Technology, Nanjing, China (gsxiehm@gmail.com).
	\IEEEcompsocthanksitem L. Wang is with New Laboratory of Pattern Recognition (NLPR), State Key Laboratory of Multimodal Artificial Intelligence Systems (MAIS), Institute of Automation, Chinese Academy of Sciences (CASIA), and also with School of Artificial Intelligence, University of Chinese Academy of Sciences (email: wangliang@nlpr.ia.ac.cn).
}}

\IEEEtitleabstractindextext{
	
\begin{abstract}
Human action understanding serves as a foundational pillar in the field of intelligent motion perception. Skeletons serve as a modality- and device-agnostic representation for human modeling, and skeleton-based action understanding has potential applications in humanoid robot control and interaction. \RED{However, existing works often lack the scalability and generalization required to handle diverse action understanding tasks. There is no skeleton foundation model that can be adapted to a wide range of action understanding tasks}. This paper presents a Unified Skeleton-based Dense Representation Learning (USDRL) framework, which serves as a foundational model for skeleton-based human action understanding. USDRL consists of a Transformer-based Dense Spatio-Temporal Encoder (DSTE), Multi-Grained Feature Decorrelation (MG-FD), and Multi-Perspective Consistency Training (MPCT). The DSTE module adopts two parallel streams to learn temporal dynamic and spatial structure features. The MG-FD module collaboratively performs feature decorrelation across temporal, spatial, and instance domains to reduce dimensional redundancy and enhance information extraction. The MPCT module employs both multi-view and multi-modal self-supervised consistency training. The former enhances the learning of high-level semantics and mitigates the impact of low-level discrepancies, while the latter effectively facilitates the learning of informative multimodal features. We perform extensive experiments on 25 benchmarks across across 9 skeleton-based action understanding tasks, covering coarse prediction, dense prediction, and transferred prediction. Our approach significantly outperforms the current state-of-the-art methods. We hope that this work would broaden the scope of research in skeleton-based action understanding and encourage more attention to dense prediction tasks. This code is available at: 
\url{https://github.com/wengwanjiang/FoundSkelModel}.
\end{abstract}

}

\maketitle

\IEEEdisplaynontitleabstractindextext
\IEEEpeerreviewmaketitle

\IEEEraisesectionheading{\section{Introduction}\label{sec:intro}}

\IEEEPARstart
Skeleton-based human action understanding has tremendous applications in areas such as robotics, human-robot interaction, immersive virtual environments, assistive technology, rehabilitation and sports analytics. With advancements in pose estimation, skeleton-based action understanding is becoming an increasingly valuable tool in AI-driven motion perception. Compared to raw video or point clouds, skeleton data is more compact, lightweight, and computationally efficient, while also offering privacy-preserving advantages. 

% With the increasing demand from applications such as human-computer interaction and intelligent surveillance, the analysis of human actions from skeleton sequences has gained substantial attention due to its robustness against background and appearance variations, as well as its privacy-preserving advantages compared to conventional RGB videos. In recent years, fully-supervised learning has achieved significant success in skeleton-based action learning~\cite{wang2018beyond,InfoGCN,zhu2023motionbert,chen2023occluded,sun2024localization}. However, these methods depend heavily on a large amount of manually annotated data. To reduce this dependence, substantial efforts have been invested in self-supervised representation learning, utilizing unlabeled data for human-centric tasks.

% fully supervised, lacks transfer
Supervised skeleton-based action recognition with deep learning has rapidly developed over the past decade. Representative methods include Hierarchical RNN~\cite{du2015hierarchical,du2016representation}, Spatio-Temporal LSTM~\cite{liu2016spatio}
Two-Stream RNN~\cite{wang2017modeling}, Multi-Modal Representations~\cite{wang2018beyond}, ST-GCN~\cite{yan2018spatial}, Two-Stream GCN~\cite{shi2019two}, Shift-GCN~\cite{cheng2020skeleton}, MS-G3D Net~\cite{liu2020disentangling} and Efficient GCN~\cite{song2022constructing}. While these approaches achieve high accuracy on the training set, they often struggle to generalize to new scenarios and unseen categories. In addition, supervised methods require a large amount of labeled data, which are time-consuming and expensive to collect. They are also prone to overfitting, especially when the dataset is small or lacks diversity. 

% self-supefvised, limited tasks
To address the above issues, self-supervised skeleton-based representation learning has gained popularity recently. Existing self-supervised approaches can be categorized into two paradigms: masked sequence modeling and contrastive learning. Masked sequence modeling leverages artificial supervision signals through pretext tasks like skeleton reconstruction~\cite{yan2023skeletonmae}, motion prediction~\cite{mao2023masked}, and skeleton colorization~\cite{yang2023self}. These methods utilize an encoder-decoder architecture to effectively capture the spatio-temporal dependencies within the skeleton sequence, enhancing the quality of representation learning. In contrast, contrastive learning-based methods~\cite{zhang2022contrastive,guo2022aimclr}, often rely on negative samples, focus on learning discriminative instance-level representations between contrastive pairs. However, both types of approaches typically require additional components such as decoders or memory banks and involve intricate masking or sampling strategies. In addition, they typically focus on learning coarse-grained action representations, neglecting the fine-grained representations, which are crucial for dense prediction tasks.

% Due to the complexity of manually crafting positive/negative pairs and their significant dependence on resources, a non-negative contrastive learning paradigm, specifically feature decorrelation, has been proposed~\cite{zhou2023self, franco2023hyperbolic}. Feature decorrelation learns distinct instance-level representations by decorrelating features and reducing redundancy across the dimensions of the feature space. However, there are still some difficulties in skeleton-based representation learning: 1) The learned representation is predominantly at the instance level, lacking the fine granularity to handle dense prediction tasks such as action detection. 2) Current feature decorrelation methods exhibit significant performance gaps compared to traditional negative-based contrastive learning approaches and lack effective training methodologies for single-modality scenarios.

\begin{figure}[t]
	\centering
	\vspace{0.4cm}
	\includegraphics[width=.9\linewidth]{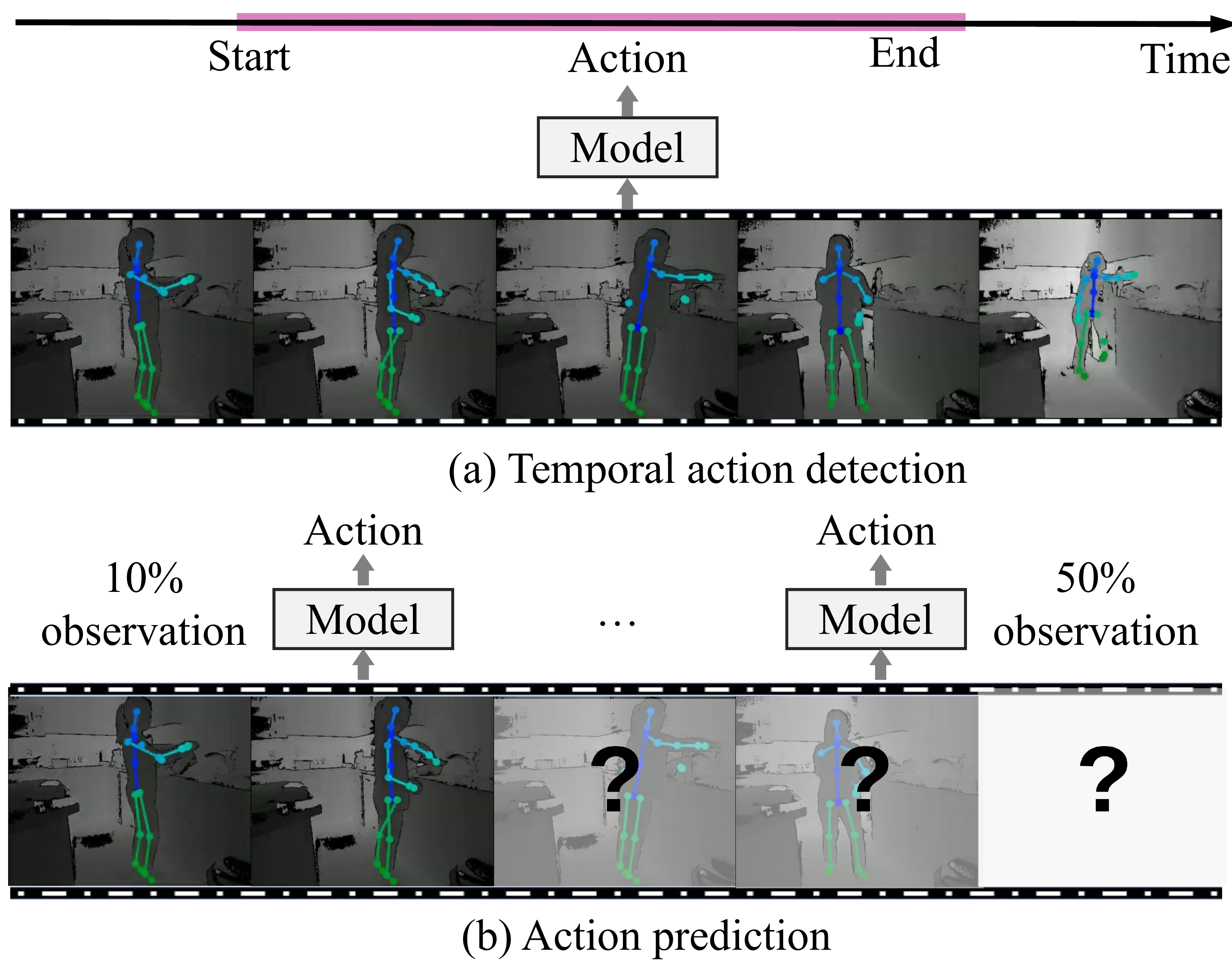}
	\caption{Dense prediction tasks of skeleton-based action understanding. Temporal action detection directly processes raw long video sequences to identify both the action categories and their temporal boundaries, while action prediction requires continuously predicting the action in an online manner and recognizing it correctly as early as possible.} 
	\label{fig: intro0}
\end{figure}

Although dense prediction tasks in image or video understanding attract significant attention, they have not received adequate attention in skeleton-based action understanding. Compared to the action recognition task which is based on pre-segmented sequences, dense prediction tasks such as temporal action detection~\cite{wang2018beyond} and action prediction~\cite{liu2018ssnet} are more aligned with real-world scenarios, as shown in Figure~\ref{fig: intro0}. 
A possible reason these problems have been overlooked is the lack of open-source benchmarks and well-recognized baselines. 
% Therefore, building a standard benchmark and baseline to promote the development of these issues is crucial.

% lack of foudation model
% key characteristics
Foundation models have recently made significant progress in image and video understanding~\cite{awais2025foundation}. For the area of skeleton-based action understanding, a foundation model should have the following characteristics: First, the model should be a simple, scalable, and easily trainable Transformer-based architecture. Second, it should be capable of learning both effective coarse-grained and fine-grained representations. Third, it should possess strong transferability across a variety of downstream tasks, including dense prediction tasks. However, most existing approaches for self-supervised skeleton-based representation learning do not meet the above requirements \RED{due to limited scalability and generalization. Dense prediction tasks, such as temporal action detection and action prediction, have been largely overlooked in existing works.}
Therefore, building a foundation model \RED{for an extensive set of tasks} of skeleton-based human action understanding is an urgent problem that needs to be addressed.

% Self-supervised skeleton-based representation learning research has mainly centered on two distinct paradigms: Masked Sequence Modeling and Contrastive Learning. Masked Sequence Modeling utilizes artificial supervision signals, incorporating mask-based pretext tasks such as skeleton reconstruction~\cite{yan2023skeletonmae}, motion prediction~\cite{mao2023masked}, and skeleton colorization~\cite{yang2023self}. By leveraging an encoder-decoder architecture, these approaches effectively model the spatio-temporal relationships within the skeleton sequence to enhance representation learning. Contrastive learning-based paradigm, which is based on negative samples~\cite{zhang2022contrastive,guo2022aimclr}, also achieves success in various downstream tasks by focusing on learning discriminative instance-level representation between contrastive pairs. However, these approaches require additional decoder or memory bank and involve sophisticated masking or sampling strategies.

\begin{figure}[t]
	\centering
	\vspace{0.4cm}
	\includegraphics[width=.9\linewidth]{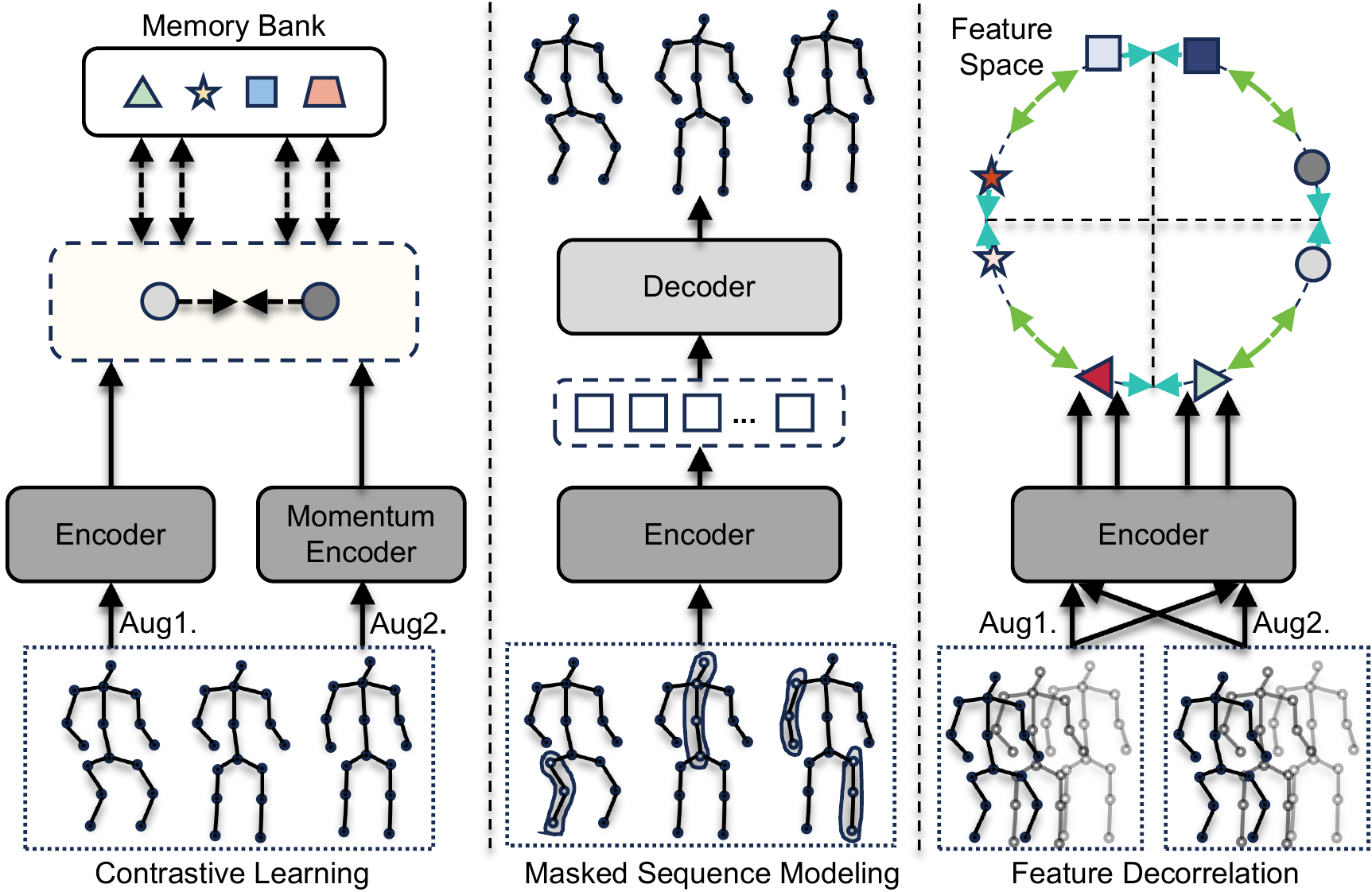}
	\caption{
		Comparison of contrastive learning-based, masked sequence modeling-based, and feature decorrelation-based self-supervised skeleton-based representation learning paradigms. The objective of the feature decorrelation-based approach is to uniformly and consistently distribute samples within the representation space. Unlike masked sequence modeling, this method is lightweight, requiring neither a decoder nor complex masking strategies. Furthermore, it streamlines and simplifies contrastive learning by eliminating the need for a memory bank or an additional momentum encoder.} 
	\label{fig: intro}
\end{figure}

To this end, we propose a foundation model with self-supervised dense representation learning named Unified Skeleton-based Dense Representation Learning (USDRL). Different from contrastive learning-based and masked sequence modeling-based methods, we use feature decorrelation for self-supervised action representation learning (see Figure~\ref{fig: intro}). 
To learn effective dense representations, we design Multi-Grained Feature Decorrelation (MG-FD) which decorrelate features across temporal, spatial, and instance domains in a multi-grained manner, ensuring both consistency within individual samples and discriminability between different samples. Additionally, we propose a Transformer-based backbone, Dense Spatio-Temporal Encoder (DSTE), which is essential for capturing multi-grained features to generate dense representations, thus enhancing the model's capacity for effective dense prediction. The DSTE comprises two modules: Convolutional Attention (CA), which captures local feature relationships, and Dense Shift Attention (DSA), which uncovers hidden dependencies. Finally, we introduce a novel Multi-Perspective Consistency Training (MPCT) framework that incorporates explicit viewpoint cues and diverse skeleton data modalities to improve representation learning while maintaining efficiency. 
We categorize skeleton-based action understanding tasks into three main types: coarse prediction, dense prediction, and transferred prediction. 
Our approach is well-suited for dense prediction tasks, including action detection and prediction.
% our method

% By decorrelating features across temporal, spatial, and instance domains in a multi-grained manner, USDRL ensures that the dense representation exhibits not only consistency within the same sample but also distinct discriminability between different samples. Furthermore, we present a Transformer-based versatile backbone called Dense spatio-temporal Encoder (DSTE), which is pivotal in capturing multi-grained features to obtain dense representations, thereby boosting the capability to perform dense prediction effectively. Specifically, the DSTE encompasses two modules: Convolutional Attention (CA) and Dense Shift Attention (DSA), which respectively model local feature relationships and uncover hidden dependencies.

% Compared to existing feature decorrelation methods, our USDRL achieves state-of-the-art performance, demonstrating the feasibility of feature decorrelation and showcasing significant advantages in skeleton-based dense representation learning. 

Our contributions can be summarized as follows:
\begin{itemize}[leftmargin=2em]
\item We present Unified Skeleton-based Dense Representation Learning (USDRL), a foundation model for skeleton-based action understanding that learns dense representations through multi-grained feature decorrelation.
\item We design a novel Dense spatio-temporal encoder to model both the temporal and spatial streams, where each stream comprises Dense Shift Attention (DSA) and Convolutional Attention (CA). The DSA captures dense dependencies, while the CA excels at integrating local features. 
\item We conduct extensive experiments on 25 benchmarks across 9 downstream action understanding tasks, including recognition, retrieval, detection and prediction, demonstrating the effectiveness of our method.
\end{itemize}

% The work is an extension of the published conference~\cite{weng2025usdrl}. Compared to the preliminary version, the major differences are as follows: First, we introduce Multi-Perspective Consistency Training (MPCT), which includes both multi-view and multi-modal self-supervised training. Second, we adapt the proposed model to three more action understanding tasks covering coarse prediction, dense prediction, and transferred prediction. 
% First, we propose a foundation model for human skeletons, and extend the USDRL to unified multimodal training to further enhance results with highly efficient early multimodal fusion. 

The work is an extension of the published conference~\cite{weng2025usdrl}. Compared to the preliminary version, the major differences are as follows: First, we propose a human skeleton-based foundation model and adapt this model to different variants for dense prediction tasks, making it well suited for over eight action understanding tasks. Second, we introduce Multi-Perspective Consistency Training (MPCT), which includes both multi-view and multi-modal self-supervised training. Third, we conduct more comprehensive experiments and analysis, including additional three experiments on action prediction, action segmentation, and transfer learning for action retrieval. 

\section{Related Work}
\label{sec_rw}
% \subsection{Self-Supervised Skeleton-Based Action Recognition}
\noindent{\textbf{Self-Supervised Skeleton-Based Action Recognition:}} 
Existing works of self-supervised skeleton representation learning can be broadly categorized into two types: masked sequence modeling and contrastive learning. Masked sequence modeling methods~\cite{yan2023skeletonmae,mao2023masked,zhu2023motionbert} learn to represent skeletons through a pretext task that involves predicting the original sequence from masked and corrupted sequences. These tasks require predicting or reconstructing the original sequence from masked and corrupted sequences, thereby capturing the spatio-temporal dynamics of the actions. 

Constrative learning-based methods can also be further subdivided into three categories: negative sample-based, self-distillation-based, and feature decorrelation-based. Negative sample-based methods~\cite{li2021crossclr,guo2022aimclr,guo2024improving} facilitate learning at the instance level by minimizing the differences with negative examples and maximizing those with positive examples within a memory bank. Self-distillation-based methods~\cite{mao2022cmd,zhang2023hierarchical} involves converting pairwise sample similarities into a probability distribution to emulate the teacher model’s probability distribution, thus capturing crucial relational knowledge. Feature decorrelation~\cite{zbontar2021barlow} is a relatively novel approach. Compared to methods using negative samples and self-distillation, these approaches~\cite{bardes2021vicreg,weng2025usdrl} are more computationally efficient and cost-effective, requiring neither large batch size nor an additional momentum encoder and memory bank.
% By decorrelating the dimensions of the representation vectors, this method enhances the learning of skeleton representations by reducing redundancy and ensuring that the dimensions of the representation vectors are as independent from each other as possible, thereby improving the quality of the representations. 

\noindent{\textbf{Skeleton-Based Action Detection/Localization:}} Temporal action detection or localization aims to identify the start time, end time, and category of each action instance in an untrimmed video. Common methods for skeleton-based action detection can be broadly categorized into two-stage and one-stage approaches. Most existing methods, such as~\cite{wang2018beyond, chen2022hierarchically, zhang2023prompted}, follow a two-stage pipeline. They first apply a frame-wise prediction approach and then generate a set of action proposals based on probabilistic matrices of action predictions for each frame. A few methods adopt a one-stage pipeline. For example, Sun et al.~\cite{sun2024localization} propose a Transformer-based end-to-end baseline that integrates deformable attention mechanisms.

% Wang et al.~\cite{wang2018beyond} combine three representations—nodes, edges, and parts—to achieve complementary advantages between different representations. 
% Song et al.~\cite{song2018spatio} propose an RNN network with spatial and temporal attention modules, which adaptively allocates different levels of attention to different frames and key joints to improve action detection performance.

\noindent{\textbf{Skeleton-Based Action Segmentation:}}
Temporal action segmentation predicts an action label to each frame of the video. Xu et al.~\cite{xu2023efficient} introduce a Connectionist Temporal Classification (CTC) loss to enhance the alignment of action segments and generate new labeled action sequences through motion interpolation. Li et al.~\cite{li2023involving} develop an improved spatial graph network to capture spatial dependencies and design a regression network to extract segmented encoding features and action boundary representations. Yang et al.~\cite{yang2023lac} focus on learning skeleton representations in both video and frame spaces, improving the model's generalization for frame-wise action segmentation. Hyder et al.~\cite{hyder2024action} use 2D skeleton heatmaps as input and apply Temporal Convolutional Networks (TCNs) to capture spatiotemporal dynamics. Ji et al.~\cite{ji2024language} incorporate language-assisted learning into skeleton-based action segmentation, utilizing linguistic knowledge to improve the understanding of relationships among joints and actions.

\noindent{\textbf{Skeleton-Based Action Prediction:}} 
Action prediction aims to recognize human actions from temporally incomplete video data. Most works focus on RGB-video based action prediction. For example, Aliakbarian et al.~\cite{aliakbarian2017encouraging} introduce a novel loss function to encourage the network to predict the correct class as early as possible. Kong et al.~\cite{kong2018action} integrate LSTM with a memory module to record the discriminative information at early stage.
% Ma et al.~\cite{ma2016learning} introduced a new ranking loss for the predictions of LSTM to enforce that either detection score of the correct activity category or the detection score margin between the correct and incorrect categories should be monotonically non-decreasing. 
Although skeleton-based action prediction is an important problem, it has not received sufficient attention. Liu et al.~\cite{liu2018ssnet} propose a network architecture based on 1D dilated convolutions to flexibly handle variations in the temporal scale of observed actions. Wang et al.~\cite{wang2019progressive} propose a teacher-student network and a cross-task progressive distillation knowledge training method to improve early-stage action prediction performance. 

% \subsection{Unified Representation Learning from Skeletons}
\noindent{\textbf{Unified Representation Learning from Skeletons:}} 
% unsupervised learning method
%Unified pretraining approaches in skeleton-based representation learning involve pretraining models on pretext tasks and then fine-tuning them on various downstream tasks to enhance the versatility and utility of the learned representations. MotionBERT~\cite{zhu2023motionbert} employs a masked sequence modeling, utilizing 2D-to-3D lifting as its pretext task and fine-tuning the entire model on various downstream tasks. Similarly, SkeletonMAE~\cite{yan2023skeletonmae} utilizes a direct reconstruction task, through masked sequence modeling to pretrain the model. PCM$^{\rm 3}$~\cite{zhang2023prompted} combines contrastive learning and masked sequence modeling to learn skeleton representations, leveraging the strengths of both methods to improve generalization across tasks. In contrast, Skeleton-in-Context~\cite{wang2024skeleton} introduces a novel In-Context Skeleton Sequence Modeling approach that captures not only skeleton positions but also the dependencies among contexts, thereby enriching the representations with environmental and intersectional features. UmURL~\cite{sun2023unified} designs its pretext task based on feature decorrelation, effectively leveraging multiple skeleton modalities. Although these methods perform well in tasks at the sequence level, they are not adept at handling dense prediction tasks at the frame level. In this work, we propose a unified approach to skeleton-based dense representation learning based on feature decorrelation, aiming to enhance the capability of capturing fine-grained features.
Unified pretraining methods in skeleton-based representation learning first train models on specific pretext tasks before adapting them to diverse downstream applications, improving their flexibility and effectiveness. MotionBERT~\cite{zhu2023motionbert} utilizes masked sequence modeling with a 2D-to-3D lifting strategy for pretraining, followed by fine-tuning on multiple tasks. SkeletonMAE~\cite{yan2023skeletonmae} employs a similar masked modeling approach but focuses on direct sequence reconstruction. PCM$^{\rm 3}$~\cite{zhang2023prompted} integrates both contrastive learning and masked modeling, leveraging their combined benefits to enhance generalization. Skeleton-in-Context~\cite{wang2024skeleton} introduces an in-context skeleton sequence modeling framework, capturing spatial relationships and contextual dependencies to refine representations. UmURL~\cite{sun2023unified} formulates its pretraining objective around feature decorrelation, effectively incorporating multiple skeleton modalities. \RED{Unified Pose Sequence (UPS)~\cite{foo2023unified} unifies heterogeneous output formats by representing both text-based action labels and coordinate-based human poses as language sequences. However, UPS targets four pose-based tasks: 3D action recognition, 2D action recognition, 3D pose estimation, and 3D early acton prediction, while our approach addresses an extensive set of action understanding tasks. Large Language Model as an Action Recognizer (LLM-AR)~\cite{qu2024llms} is also proposed; however, LLM-AR focuses solely on the action recognition task.
}
Wang et al.~\cite{wang2025heterogeneous} investigate data heterogeneity in human skeletons and propose a unified framework for learning action representation from different  heterogeneous skeletons. Although these techniques perform well in sequence-level tasks, they struggle with frame-level dense prediction. To bridge this gap, we propose a novel skeleton-based dense representation learning framework centered on feature decorrelation, aiming to improve fine-grained feature capture.

% subsection{Feature Decorrelation}
\noindent{\textbf{Feature Decorrelation:}} 
% Feature decorrelation is a method used to prevent model collapse in self-supervised learning by decorrelating features and reducing redundancy across the dimensions of the learned features. Before this, a common method to prevent model collapse is negative-based contrastive learning. However, this approach typically requires an additional momentum encoder, a large memory bank, and a significant batch size. To address this issue, W-MSE~\cite{ermolov2021whitening} introduces a non-negative based method (i.e., feature decorrelation). In W-MSE, a pair of positive samples (\textit{e.g.}, different augmentations of the same image) are encoded using a shared encoder. After encoding, a whitening module is added to whiten all embeddings within each batch, ensuring the variables of each dimension in embeddings are linearly independent. Barlow Twins~\cite{zbontar2021barlow} employs a similar approach to W-MSE, but it calculates the cross-correlation matrix between the variables of the two vectors and optimizes this matrix to be close to an Identity, aiming for the same goal as W-MSE. Inspired by Barlow Twins, VICREG~\cite{bardes2021vicreg} constrains the self-supervised learning process by combining three distinct types of losses: variance, invariance, and covariance, and achieves stabilization without any normalization to prevent model collapse. In this work, we integrate the Barlow Twins and VICREG methods to conduct self-supervised skeleton-based representation learning based on feature decorrelation.
Feature decorrelation is a technique in self-supervised learning designed to prevent model collapse by minimizing redundancy across feature dimensions and ensuring diverse representations. Traditionally, contrastive learning with negative samples has been used to tackle this issue, but it often requires additional components such as a momentum encoder, a large memory bank, and substantial batch sizes, making it resource-intensive. To provide an alternative, W-MSE~\cite{ermolov2021whitening} introduces a decorrelation-based approach that does not rely on negative samples. In W-MSE, positive sample pairs (\textit{e.g.}, different augmentations of the same image) are processed through a shared encoder, followed by a whitening module that standardizes feature distributions and ensures linear independence across dimensions. Barlow Twins~\cite{zbontar2021barlow} employs a related strategy but optimizes the cross-correlation matrix between two encoded representations, encouraging it to approximate an identity matrix to achieve the same goal. Building on this idea, VICREG~\cite{bardes2021vicreg} stabilizes self-supervised learning by incorporating variance, invariance, and covariance to prevent collapse. In this work, we integrate Barlow Twins and VICREG principles to develop a self-supervised skeleton-based representation learning framework using feature decorrelation.

\noindent{\textbf{Video Foundation Models:}} Video Foundation Models aim to develop a general-purpose representation for diverse video understanding tasks. Our review focuses solely on works that achieve effective representation learning through pre-training with specific loss functions, broadly classified into discriminative and generative approaches. Typical discriminative pretraining are Video-Text Contrastive~\cite{xu2021videoclip} and Video-Text Matching~\cite{li2022align}. The former pulls similar representations together while pushing dissimilar ones apart, whereas the latter aims to maximize the matching score for a given video-text pair. Generative approaches aim to reconstruct masked information within video data. Notable examples include Masked Language Modeling~\cite{xue2022advancing}, Mask Video Modeling~\cite{wang2023videomae}, and Mask Image Modeling~\cite{wang2022bevt}.

\noindent{\textbf{\RED{Foundational Motion Models:}} } \RED{There are some foundational motion models that unify human motion prediction and synthesis, trajectory prediction, 3D pose estimation, and action understanding, e.g., Unified Pose Sequence (UPS)\cite{foo2023unified}, SkeletonMAE\cite{yan2023skeletonmae}, MotionBERT~\cite{zhu2023motionbert}, Sports-Traj~\cite{xusports} and Multi-modal Large Motion Model~\cite{zhang2024large}. While motion synthesis emphasizes generating new motions consistent with semantic input, motion understanding focuses on interpreting and analyzing motions. 
}

\begin{figure*}[tb]
	\centering
	\includegraphics[width=0.98\linewidth]{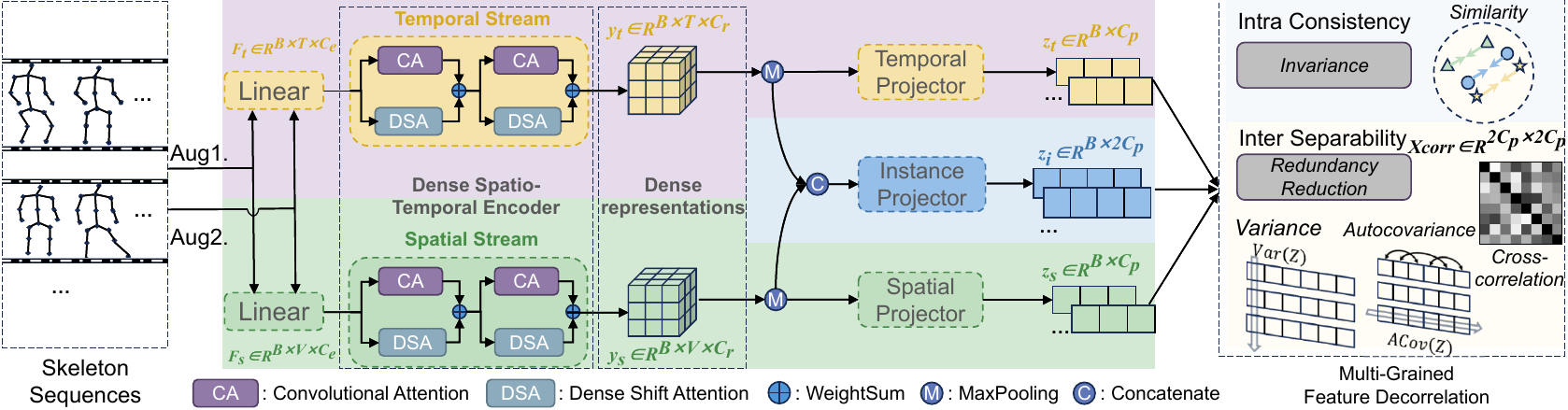}
	\caption{
		The proposed Unified Skeleton-based Dense Representation Learning (USDRL) framework. USDRL incorporates a two-stream architecture with the Dense Spatio-Temporal Encoder (DSTE). The DSTE processes skeleton sequences to derive dense representations, which are further refined through MaxPooling and concatenation to generate condensed vectors.
		The Multi-Grained Feature Decorrelation is devised to mitigate model collapse and guarantee both intra-sample consistency and inter-sample separability. To further enhance robustness across different viewpoints and facilitate multimodal learning, a multi-perspective consistency training strategy is used during training.
		% Utilizing vectors from the projectors, we prevent model collapse by calculating multi-grained feature decorrelation, emphasizing Intra-sample Consistency and Inter-sample Separability. Beyond conventional similarity metrics, we compute autocorrelation along the channel dimension and variance across batches. Notably, the cross-correlation is split along the diagonal, highlighted in red parts.
	} 
% and three domain-specific projectors
	\label{fig:framwork}
\end{figure*}

\section{Methods}
\label{sec_method}

We introduce a foundation model for skeleton-based human action understanding. As depicted in Figure \ref{fig:framwork}, this model follows the standard two-stream pipeline~\cite{wang2017modeling} and primarily consists of a Dense Spatio-Temporal Encoder (DSTE). During training, three specialized projectors, namely temporal, spatial, and instance projectors, each composed of multiple linear layers, are employed. Additionally, a self-supervised training paradigm called Multi-Grained Feature Decorrelation (MG-FD) is proposed. Unlike contrastive methods relying on negative samples~\cite{guo2022aimclr}, this technique eliminates the need for a momentum encoder or memory bank.

During the forward pass, an augmented 3D skeleton sequence $\mathbf{X} \in \mathbb{R}^{C_{in} \times T \times V \times M}$ of length $T$ with $V$ skeletal vertices is first transformed into two distinct views: temporal and spatial. The sequence is reshaped into $\mathbf{X}_{t} \in \mathbb{R}^{T \times (M \times V \times C_{in})}$ for the temporal domain and $\mathbf{X}_{s} \in \mathbb{R}^{(M \times V) \times (T \times C_{in})}$ for the spatial domain. These rearranged sequences are then mapped into an embedding space, producing feature representations $\mathbf{F}_{t} \in \mathbb{R}^{T \times C_e}$ and $\mathbf{F}_{s} \in \mathbb{R}^{V \times C_e}$. Next, these embeddings pass through the DSTE’s respective temporal and spatial streams, generating dense representations $\mathbf{y}_{t} \in \mathbb{R}^{T \times C_r}$ and $\mathbf{y}_{s} \in \mathbb{R}^{V \times C_r}$. Applying MaxPooling condenses these into vectors $\mathbf{y}_{t}, \mathbf{y}_{s} \in \mathbb{R}^{C_r}$. These are subsequently projected into a new space using the three domain-specific projectors, yielding projection vectors $\mathbf{z}_{t}, \mathbf{z}_{s} \in \mathbb{R}^{C_p}$ and $\mathbf{z}_{i} \in \mathbb{R}^{2 \times C_p}$. Here, $C_{in}, C_e, C_r, C_p$ represent the channel dimensions for input, embedding, representation, and projection, respectively. Finally, the proposed MG-FD, $\mathcal{L}_{mfd}$, is computed among the projection vectors to enforce decorrelation constraints across temporal, spatial, and instance features.

\subsection{Dense Spatio-Temporal Encoder}
\label{sec_encoder}

% Our Dense Spatio-Temporal Encoder comprises a temporal stream and a spatial stream, each modeling the temporal and spatial dimensions, respectively. Both streams are composed of multiple stacked layers, as shown in Figure \ref{fig: encoder}. Each layer contains a Dense Shift Attention (DSA) Module and a Convolutional Attention (CA) Module. The detailed structure of the DSTE layer is described below.

\begin{figure}[tb]
\centering
\includegraphics[width=.95\linewidth]{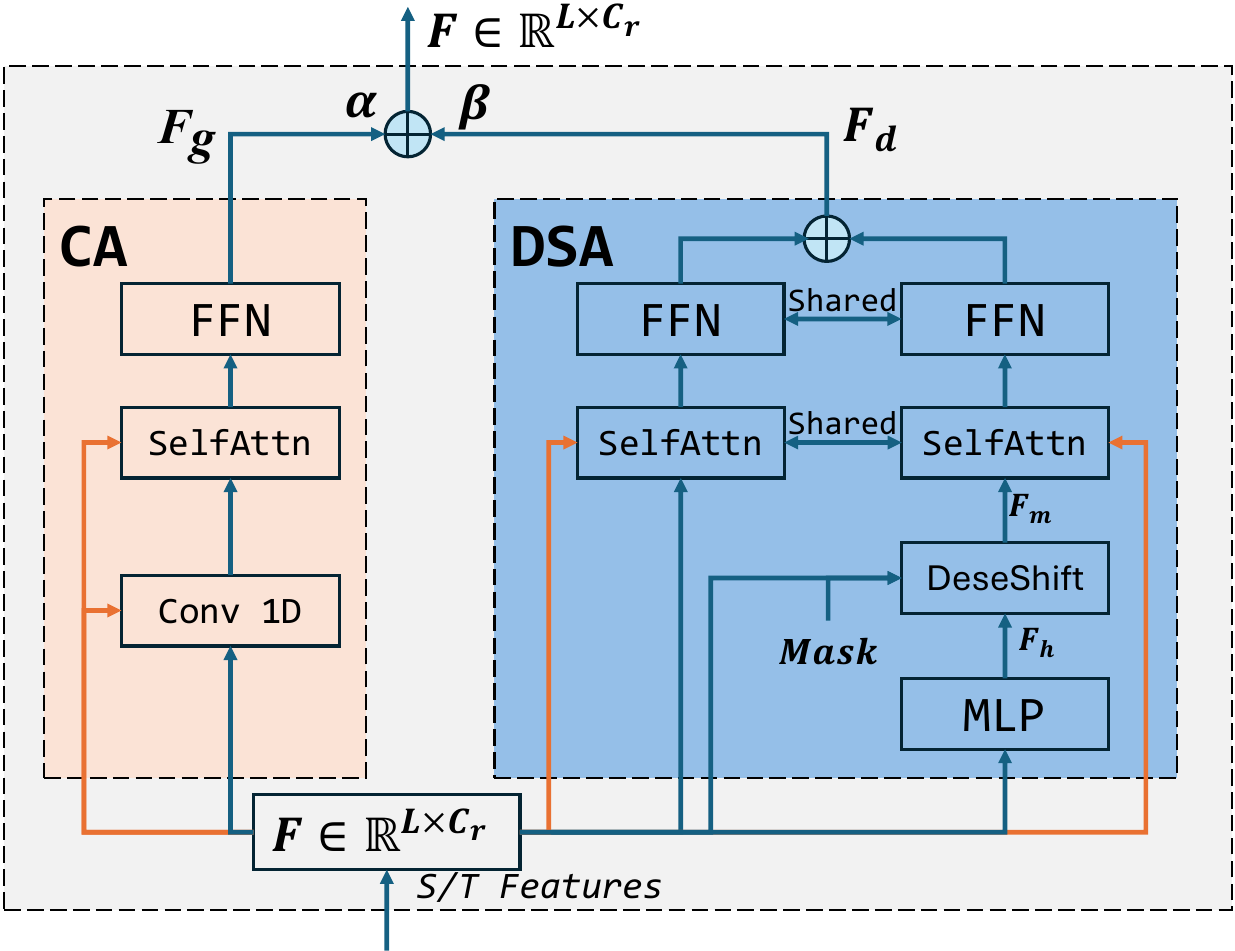}
	\caption{The fundamental layer of the Dense Spatio-Temporal Encoder. It consists of the Convolutional Attention (CA) and Dense Shift Attention (DSA) modules, where $\oplus$ represents a weighted sum operation.}  
\label{fig: encoder}
\end{figure}

The proposed Dense Spatio-Temporal Encoder (DSTE) consists of two parallel temporal and spatial streams designed to capture dynamic and structural features, respectively. As illustrated in Figure \ref{fig: encoder}, both streams are constructed using multiple stacked layers, each integrating a Dense Shift Attention (DSA) module and a Convolutional Attention (CA) module. The architecture of a single DSTE layer is detailed below.  

\noindent{\textbf{Dense Shift Attention (DSA):} }Given an input embedding sequence $\mathbf{F} \in \mathbb{R}^{L \times C_e}$ in either the temporal or spatial domain, the DSA module leverages an MLP with two learnable weight matrices, $W_{1}, W_{2} \in \mathbb{R}^{L \times L}$, to uncover underlying relationships between embeddings. Here, $L$ represents the sequence length. Specifically, the sequence is first reshaped into $\mathbf{F}_{1} \in \mathbb{R}^{C_e \times L}$, then processed as follows:  
% \paragraph{Dense Shift Attention (DSA):} Given the temporal or spatial embeddings sequence $\mathbf{F} \in \mathbb{R}^{L \times C_e}$, the DSA employs an MLP composed of two learnable weight matrices $W_{1}, W_{2} \in \mathbb{R}^{L \times L}$ to reveal hidden relationships among all embeddings within the sequence, where $L$ denotes the length of $\mathbf{F}$. Specifically, we apply an MLP to $\mathbf{F}_{1} \in \mathbb{R}^{C_e \times L}$, which is reshaped from the embeddings sequence $\mathbf{F}$:
\begin{equation}\label{DSA_D}
    \mathbf{F}_{h} = \text{ReLU} \left(W_{1}\mathbf{F}_{1}\right)W_{2} + \mathbf{F}_{1}.
\end{equation}

The transformed representation $\mathbf{F}_{h}$, enriched with global context, is then combined with the original sequence $\mathbf{F}$ through a DenseShift operation. This mechanism enables embeddings to incorporate semantic information from the entire sequence:  
% Subsequently, $\mathbf{F}_{h}$, enriched with global information, is fused with the original sequence $\mathbf{F}$  via a DenseShift operation. This operation facilitates each embedding to assimilate semantic information from across the entire sequence. The DenseShift operation is formally defined as follows:
% The DenseShift operation between $\mathbf{F}_{h}$ and $\mathbf{F}$ is defined in Eq. \ref{gap}, detailing how $\mathbf{F}_{m}$ is derived.
\begin{equation}\label{gap}
    \mathbf{F}_{m} =\mathbf{Mask} \odot \mathbf{F}_{h}  + \overline{\mathbf{Mask}} \odot \mathbf{F},
\end{equation}
% where $\mathbf{Mask}$ is a binary vector where every $gap$-th element is set to 1 and all other elements are set to 0, and $\overline{\mathbf{Mask}}$ is the inverse of $\mathbf{Mask}$, i.e, $\overline{\mathbf{Mask}} = 1 - \mathbf{Mask}$. 
where $\mathbf{Mask}$ is a binary vector with every $gap$-th element set to 1 and all others set to 0, while $\overline{\mathbf{Mask}} = 1 - \mathbf{Mask}$.  

Subsequently, $\mathbf{F}_{m}$ and $\mathbf{F}$ undergo sparse Self-Attention (SA) and Feed-Forward Network (FFN) transformations. The final output of the DSA module, $\mathbf{F}_{d}$, is obtained via: 
\begin{equation}\label{DSA_SA}
    \mathbf{F}_{d} = \operatorname{FFN}\left(\operatorname{SA}\left(\mathbf{F}_{m}\right)\right) + \operatorname{FFN}\left(\operatorname{SA}\left(\mathbf{F}\right)\right),
\end{equation}
where $\mathbf{F}_{d}$ is the output of the DSA module. 
% This module is partly inspired by dense temporal modeling across both spatial and channel domains~\cite{wang2017modeling,xing2023boosting}.

\noindent{\textbf{Convolutional Attention (CA):}} 
% The CA module begins applying 1D channel-wise temporal/spatial convolution operations on the embeddings $\mathbf{F} \in \mathbb{R}^{T/V \times C_e}$, enhancing feature interactions within the sequence. It subsequently employs self-attention operations on these convolved features to capture long-term dependencies and model global interactions, thereby producing $\mathbf{F_g} \in \mathbb{R}^{T/V \times C_r}$. This operation is defined as follows:
The CA module begins by applying 1D channel-wise temporal or spatial convolutions on the input embeddings $\mathbf{F} \in \mathbb{R}^{T/V \times C_e}$ to enhance local feature interactions. The resulting representations then pass through a self-attention mechanism, capturing long-range dependencies and global patterns. The final output $\mathbf{F}_{g} \in \mathbb{R}^{T/V \times C_r}$ is computed as follows:  
\begin{equation}\label{laa}
    \mathbf{F}_{g} = \operatorname{FFN}\left( \operatorname{SA}\left(\operatorname{Conv}\left(\mathbf{F}\right)
    + \mathbf{F} \right) \right),
\end{equation} 
where $\operatorname{FFN}, \operatorname{SA}$ denote Feed-Forward layer and Self-Attention layer, respectively.

\noindent{\textbf{Spatio-Temporal Representations:}}
% The outputs from the CA and DSA are combined through a weighted sum:
The outputs of the CA and DSA modules are combined using a weighted sum:  
\begin{equation}\label{DSA}
    \mathbf{y} =\alpha \operatorname{CA}\left(\mathbf{F}\right) + \beta \operatorname{DSA}\left(\mathbf{F}\right),
\end{equation}
% where $\alpha$ and $\beta$ are the weight coefficients, $\alpha + \beta = 1$, and $\operatorname{DSA}(\cdot)$ and $\operatorname{CA}(\cdot)$ denote the functions defined in Eq.~\ref{DSA_SA} and Eq.~\ref{laa}, respectively.  
where $\alpha$ and $\beta$ are weighting coefficients satisfying $\alpha + \beta = 1$. The functions $\operatorname{DSA}(\cdot)$ and $\operatorname{CA}(\cdot)$ refer to the operations defined in Eq. \ref{DSA_SA} and Eq.~\ref{laa}, respectively.  
% using the hyperparameters $\alpha$ and $\beta$, as defined in Eq. \ref{DSA}, to obtain the dense temporal/spatial representation $\mathbf{y}$.

% The outputs from the temporal and spatial streams, $\mathbf{y}_{t} \in \mathbb{R}^{T \times C_{r}}$ and $\mathbf{y}_{s} \in \mathbb{R}^{V \times C_{r}}$, respectively, serve as the dense temporal and spatial representations:
The final dense spatio-temporal representations, $\mathbf{y}_{t} \in \mathbb{R}^{T \times C_{r}}$ and $\mathbf{y}_{s} \in \mathbb{R}^{V \times C_{r}}$, serve as the core feature representations:  
\begin{equation}\label{layer}
    \mathbf{y}_s, \mathbf{y}_t = \operatorname{DSTE} \left( \mathbf{F_s},  \mathbf{F_t}\right),
\end{equation}
% where $\operatorname{DSTE}(\cdot)$ denotes the Dense Spatio-Temporal Encoder. These action representations preserve temporal and spatial dimensions of the skeleton sequence.
where $\operatorname{DSTE}(\cdot)$ denotes the Dense Spatio-Temporal Encoder. These representations effectively preserve both the temporal dynamics and spatial structure of the skeleton sequence.  

\subsection{Multi-Grained Feature Decorrelation}
\label{sec_loss}

After mapping the learned action representations to a higher-dimensional space via three specialized projectors, a Multi-Grained Feature Decorrelation training loss is formulated, specifically designed for the temporal, spatial, and instance domains. The overall training loss \(\mathcal{L}\) is expressed as:

\begin{equation}\label{total_loss}
	\mathcal{L} = \mathcal{L}_{fd} \left(\mathbf{Z}\right) +
	\tau \left( \mathcal{L}_{fd} \left(\mathbf{Z}_{s}\right) +
	\mathcal{L}_{fd} \left(\mathbf{Z}_{t}\right) \right),
\end{equation}
where \(\mathbf{Z_s}\), \(\mathbf{Z_t}\), and \(\mathbf{Z}\) represent feature matrices containing high-dimensional representations for spatial, temporal, and instance domains, respectively. Here, \(\mathcal{L}_{fd}\) signifies the feature decorrelation loss, and \(\tau\) is a hyperparameter that regulates the contributions of spatial and temporal components.

The feature decorrelation loss incorporates both Intra-Sample Consistency and Inter-Sample Separability. We elaborate on these components in the context of the instance domain as an example.

\noindent{\textbf{Intra-Sample Consistency:}} Since data augmentation preserves intrinsic information, projection vectors \(\mathbf{z}_k\) derived from \(K\) augmentations should retain semantic consistency. To enforce this, we introduce \(\mathcal{L}_{con}\), comprising a \textit{Similarity term} and an \textit{Invariance term}. The Similarity term minimizes the mean squared error (MSE) among augmentations, ensuring close proximity of representations, whereas the Invariance term aligns their autocorrelation towards 1, promoting consistency. \RED{\(\mathcal{L}_{con}\) is defined as:}
\begin{equation}\label{mse}
	\mathcal{L}_{con} = \frac{1}{K} \sum_{a=1}^{K}\Big(\kappa \underbrace{ \left \| \mathbf{z}_a - \overline{\mathbf{z}} \right \| _{2}}_{\text{Similarity}}  +
	\eta \sum_{b=1|b \neq a}^{K}  \underbrace{\operatorname{tr}\left(\mathbf{I} - \widehat{\mathbf{z}}_a^{\operatorname{T}} \widehat{\mathbf{z}}_b \right)}_{\text{Invariance}} \Big),
\end{equation}
where \(\widehat{\mathbf{z}}\) is the normalized vector, \(\overline{\mathbf{z}}\) is the average across augmentations, \(\operatorname{tr}\) is the trace operator, \(\mathbf{I}\) denotes the identity matrix, and \(\eta, \kappa\) are weighting factors.

\noindent{\textbf{Inter-Sample Separability:}} Without inter-sample separability, representations may become highly correlated, leading to redundancy and model collapse. Inspired by recent self-supervised learning advancements, we define the Inter-Sample Separability loss \(\mathcal{L}_{sep}\) using three complementary terms: variance, covariance, and cross-correlation, which together ensure feature decorrelation and distinguishability across samples.

\textit{Variance Term.} Given a matrix \(\mathbf{Z} \in \mathbb{R}^{N \times C_p}\), where \(N\) is the batch size and \(C_p\) represents feature dimensions, the variance term maintains feature diversity by ensuring each dimension has a variance exceeding a threshold \(\gamma\). A small constant \(\epsilon\) prevents instability:
\begin{equation}\label{vac_v}
	V\left(\mathbf{Z}\right) =\frac{1}{C_p} \sum_{j=1}^{C_p} \text{ReLU}\Big(\gamma - \sqrt{Var\left(\mathbf{Z}_{:,j}\right)+\epsilon} \Big),
\end{equation}
where \(Var(\mathbf{Z}_{:,j})\) denotes variance along dimension \(j\).

\textit{Auto-Covariance Term.} To promote independence among features, the auto-covariance term minimizes cross-dimension correlations within \(\mathbf{Z}\):
\begin{equation}\label{vac_a}
	AC\left(\mathbf{Z}\right)= \frac{1}{C_p}\sum_{i=1}^{C_p} {\sum_{j=1 | j \neq i}^{C_p}{\left[ACov\left(\mathbf{Z}\right)\right]_{i,j}^{2}}},
\end{equation}
where \(ACov(\mathbf{Z})\) is the auto-covariance matrix.

\textit{Cross-Correlation (Xcorr) Term.} This term reduces redundancy by decorrelating features among different augmentation versions. Given two augmented feature sets \(\mathbf{Z}_a\) and \(\mathbf{Z}_b\), we compute their cross-correlation matrix. The off-diagonal elements should ideally be close to zero, ensuring minimal redundancy:
\begin{equation}\label{vac_c}
	XC\left(\mathbf{Z}_a,\mathbf{Z}_b\right)= \sum_{i=1}^{C_p} {\sum_{j=1|j \neq i}^{C_p} \left[Xcorr\left(\mathbf{Z}_{a},\mathbf{Z}_{b}\right)\right]_{i,j}^{2}},  
\end{equation}
where \(Xcorr(\mathbf{Z}_a,\mathbf{Z}_b)\) is the cross-correlation matrix.

By integrating these terms, the final Inter-Sample Separability loss \(\mathcal{L}_{sep}\) is formulated as:
\begin{equation}\label{vac}
	\mathcal{L}_{sep} = \sum_{a=1}^{K} {\Big(\mu V\left(\mathbf{Z}_a\right) +  AC\left(\mathbf{Z}_a\right) +
		\lambda \sum_{b=a+1}^{K} {XC\left(\mathbf{Z}_a, \mathbf{Z}_b\right)} \Big)},
\end{equation}
where \(\mu\) and \(\lambda\) balance the contributions of different terms. Finally, the instance domain loss is given by:
\begin{equation}\label{instance_loss}
	\mathcal{L}_{fd}\left(\mathbf{Z}\right) = \mathcal{L}_{con}\left(\mathbf{Z}\right) + \mathcal{L}_{sep}\left(\mathbf{Z}\right).
\end{equation}

\subsection{Multi-Perspective Consistency Training}
\noindent{\textbf{Multi-View Training:}} For self-supervised training, most works treat different data-augmented versions of the same sample as positive samples to learn augmentation-invariant features. As acquiring multiple viewpoints of the same action sequence is relatively easy, we explicitly incorporate viewpoint information for self-supervised action representation learning. 
% However, obtaining sets of different viewpoints for the same action sequence is relatively simple. This can be accomplished by setting up multiple cameras to record simultaneously, providing easy access to different viewpoints without the need for manual annotation or supervision. 
Positive pairs are constructed not only via data augmentation, but also from the same action sequence captured from different viewpoints. This strategy encourages the model to learn representations that are both augmentation-invariant and viewpoint-invariant. Specifically, inputs for the unified and single-modality branches are derived from the same action sequence recorded by different cameras, with each view subjected to distinct data augmentations. Although the single-modality inputs originate from the same underlying sequence, they undergo separate augmentations to enhance input diversity and reduce reliance on low-level cues during inter-modal alignment. Through this approach, the model is guided to focus on high-level semantic features of the action itself, while mitigating the influence of low-level discrepancies caused by viewpoint variations.

\noindent{\textbf{Multi-Modal Training:}} Different modalities~\cite{wang2018beyond}, such as joints, bones, and motion, can be used to represent a given skeleton sequence. Most existing works combine multimodal skeleton results using late fusion, which employs three separate streams and increases the computational cost by three times. 
Let $\mathbf{X^k}$ represent a specific modality, where $k$ is the index of the various modalities, i.e., $k \in \{joint, bone, motion\}$. The input from different modalities is mapped to a common high-dimensional embedding space through distinct linear embedding modules to obtain $\mathbf{F^k}$. To reduce computational complexity, early fusion is employed to integrate embeddings from different modalities into a unified representation. For both the spatial and temporal streams, the multimodal dense representation is computed as:
% This unified embedding is then fed into the skeleton encoder to obtain the fused action representation. These steps can be summarized as follows:
\begin{equation}
	\label{fusion}
	\mathbf{y} = \mathrm{{DSTE}}(\mathrm{{Fusion}}(\mathbf{F}^{joint}, \mathbf{F}^{bone}, \mathbf{F}^{motion})),
\end{equation}
where $\mathbf{y}$ denotes the fused dense representation of joint, bone, and motion modalities, and $\mathrm{{Fusion}}(\cdot)$ denotes the early fusion operations, such as averaging.

For the purposes of simplicity and efficiency, our approach omits the inter-modal consistency loss when compared with UmURL~\cite{sun2023unified}. As multimodal features share the common encoder backbone, early fusion significantly reduces model parameters in comparison to other approaches that maintain separate backbones for multimodal inputs. 
% Our findings indicate that this straightforward alignment of different modalities is ineffective and may even degrade performance.

\subsection{Adaptation for Dense Prediction Tasks}
\noindent{\textbf{Action Detection:}} For temporal action detection, we follow the two-stage pipeline~\cite{wang2018beyond} that considers action detection as a frame-wise classification problem, with an additional action class representing the background.  During inference, for long videos, we adopt a sliding window approach to sample fixed-length segments, feed them into the network, and then concatenate the predictions along the temporal dimension. A post-processing step is used to transform the predicted frame-wise probabilities into a set of triplets, each consisting of a start frame, an end frame, and an action category. 

\noindent{\textbf{Action Segmentation:}} Skeleton-based action segmentation aims to predict the action category for each frame. We follow the end-to-end pipeline~\cite{li2023decoupled} that directly feeds the long sequence into the network.

\noindent{\textbf{Action Prediction:}} Action prediction aims to recognize the action as early as possible given a partially observed action sequence. For such task that requires temporal causality, we employ causal attention operations in the transformer for the temporal stream. 

\section{Experiment}
\label{sec_exp}

\subsection{Datasets} 
NTU-RGB+D 60 Dataset (NTU-60)~\cite{shahroudy2016ntu} is a dataset for skeleton-based human action recognition, featuring 56,880 sequences, where each sequence represents a single action performed by one of 40 subjects across 60 categories. Each action is captured as 3D coordinates of 25 body joints. We employ the established Cross-Subject (X-sub) and Cross-View (X-view) evaluation protocols. 
% For the X-sub protocol, training is conducted on 20 subjects, while testing is performed on the remaining 20 subjects. For the X-view protocol, training utilizes data from cameras 2 and 3, and testing is subsequently carried out using data from camera 1.

NTU-RGB+D 120 Dataset (NTU-120)~\cite{liu2019ntu} is an extended version of NTU-60, featuring 114,480 sequences across 120 action categories performed by 106 subjects. Each action, represented by 3D coordinates of 25 body joints, is evaluated using Cross-Subject (X-sub) and Cross-Setup (X-set) protocols. In the X-sub, half of the subjects are used for training and the rest for testing. The X-set splits the data from 32 camera setups for diverse training and testing.

PKU Multi-Modality Dataset (PKUMMD)~\cite{liu2020benchmark} is a large-scale, multi-modality dataset for 3D human action understanding, with two subsets: I and II. PKUMMD I is dedicated to action detection, using continuous video streams to identify and recognize specific actions. PKUMMD II focuses on tasks of action recognition and segmentation. The recognition crops action instances based on temporal annotations and aims to classify them into 51 distinct categories using a cross-subject protocol, while the segmentation aims to directly segment the raw skeleton sequence. This dataset presents significant challenges due to variations in camera angles and subject positioning, comprising 5,332 training samples and 1,613 testing samples.

\RED{UAV-Human~\cite{li2021uav} is a large-scale human action understanding dataset captured by unmanned aerial vehicles. We use the 2D skeleton sequences for action recognition.}

\begin{table*}[t]
	\centering
	\caption{Comparison of unsupervised action recognition results. J: Joint, M: Motion, B: Bone.}
	  % \vspace{-0.05in}
	% \addtolength{\tabcolsep}{-4.0pt}
	% \resizebox{0.84\textwidth}{!}{ 
		\begin{tabular}{llccccccc}
			\toprule
			\multirow{2}[4]{*}{\textbf{Method}} & \multirow{2}[4]{*}{\textbf{Publisher}} & \multirow{2}[4]{*}{\textbf{Modality}} & \multicolumn{2}{c}{\textbf{NTU-60}} & \multicolumn{2}{c}{\textbf{NTU-120}} & \multicolumn{2}{c}{\textbf{PKU-MMD II}} \\
			\cmidrule{4-9}          &       &       & x-sub  & x-view & x-sub  & x-set  & \multicolumn{2}{c}{x-sub} \\
			% \midrule
			% MS$^\text{2}$L~\cite{MS2L} & ACM MM'20 & J     & 52.6  & -     & -     & -     & \multicolumn{2}{c}{27.6} \\
			% ViA~\cite{yang2024view}  & IJCV'24& J+M     & 78.1  & 85.8  & 69.2    & 66.9  & \multicolumn{2}{c}{-} \\
			% PCM$^{\rm 3}$~\cite{zhang2023prompted}  & ACM MM'23 & J     & 83.9  & 90.4  & 76.5  & 77.5  & \multicolumn{2}{c}{51.5} \\
			% \midrule
			% P\&C~\cite{su2020predict}  & CVPR'20 & J     & 50.7  & 76.1  & -  & -  & \multicolumn{2}{c}{-} \\
			\textit{\textbf{Masked Sequence Modeling}} &       &       &       &       &       &       & \multicolumn{2}{c}{} \\
			% LongT GAN~\cite{zheng2018unsupervised} & AAAI'18 & J     & 52.1  & 56.4  & -     & -     & \multicolumn{2}{c}{26.5} \\
			MS$^\text{2}$L~\cite{MS2L} & ACM MM'20 & J     & 52.6  & -     & -     & -     & \multicolumn{2}{c}{27.6} \\
			3s-Colorization~\cite{yang2021skeleton}  & ICCV'21 & J     & 75.2  & 83.1  & -  & -  & \multicolumn{2}{c}{-} \\
			GL-Transformer~\cite{kim2022global} & ECCV'22 & J     & 76.3  & 83.8  & 66    & 68.7  & \multicolumn{2}{c}{-} \\
			Masked Colorization~\cite{yang2023self} & TPAMI'23 & J     & 79.1  & 87.2  & 69.2  & 70.8  & \multicolumn{2}{c}{49.8} \\
			PCM$^{\rm 3}$~\cite{zhang2023prompted}  & ACM MM'23 & J     & 83.9  & 90.4  & 76.5  & 77.5  & \multicolumn{2}{c}{51.5} \\
			SkeletonMAE~\cite{wu2023skeletonmae}  & ICMEW'23 & J     & 74.8  & 77.7  & 72.5  & 73.5  & \multicolumn{2}{c}{36.1} \\
			MAMP~\cite{mao2023masked}  & ICCV'23 & J     & 84.9  & 89.1  & 78.6  & 79.1  & \multicolumn{2}{c}{53.8} \\
			MacDiff~\cite{wu2024macdiff}  & ECCV'24 & J     & 86.4  & 91.0  & 79.4  & 80.2  & \multicolumn{2}{c}{-} \\
			MMFR~\cite{zhu2024motion}  & TCSVT'24 & J     & 84.2  & 89.5  & 77.1  & 78.8  & \multicolumn{2}{c}{54.4} \\
			\midrule
			\textit{\textbf{Contrastive Learning}} &       &       &       &       &       &       & \multicolumn{2}{c}{} \\
			
			AimCLR~\cite{guo2022aimclr} & AAAI'22 & J     & 74.3  & 79.7  & 63.4  & 63.4  & \multicolumn{2}{c}{38,5} \\
			CMD~\cite{mao2022cmd}   & ECCV'22 & J     & 79.8  & 86.9  & 70.3  & 71.5  & \multicolumn{2}{c}{43.0}  \\
			CPM~\cite{zhang2022contrastive}   & ECCV'22 & J     & 78.7  & 84.9  & 68.7  & 69.6  & \multicolumn{2}{c}{48.3}  \\
			PSTL~\cite{zhou2023self}  & AAAI'23 & J     & 77.3  & 81.8  & 66.2  & 67.7  & \multicolumn{2}{c}{49.3} \\
			HaLP~\cite{Shah_2023_CVPR}  & CVPR'23 & J     & 79.7  & 86.8  & 71.1  & 72.2  & \multicolumn{2}{c}{43.5} \\
			HiCo~\cite{hico2023}  & AAAI'23 & J     & 81.1  & 88.6  & 72.8  & 74.1  & \multicolumn{2}{c}{49.4} \\
			2s-DMMG~\cite{guan2023dmmg}  & TIP'23 & J+M     & 84.2  & 89.3  & 72.7  & 72.4  & \multicolumn{2}{c}{58.8} \\
			Skeleton-logoCLR~\cite{hu2024global}  & TCSVT'24 & J     & 82.4  & 87.2  & 72.8  & 73.5  & \multicolumn{2}{c}{54.7} \\
			KTCL~\cite{wang2024localized}  & TMM'24 & J     & 82.4  & 89.4  & 74.4  & 74.5  & \multicolumn{2}{c}{55.5} \\
			SCD-Net~\cite{wu2024scd}  & AAAI'24 & J     & 86.6  & 91.7  & 76.9  & 80.1  & \multicolumn{2}{c}{54.0} \\
			ViA~\cite{yang2024view}  & IJCV'24& J+M     & 78.1  & 85.8  & 69.2    & 66.9  & \multicolumn{2}{c}{-} \\
			IGM~\cite{lin2024idempotent}   & ECCV'24 & J     & 86.2  & 91.2  & 80.0  & 81.4  & \multicolumn{2}{c}{-}  \\
			% 4s-MG-AL~\cite{yang2022motion}  & TCSVT'22 & J+M    & 64.7  & 68.0  & 46.2  & 49.5  & \multicolumn{2}{c}{-} \\
			\midrule
			\textit{\textbf{Feature Decorrelation}} &       &       &       &       &       &       & \multicolumn{2}{c}{} \\
			HYSP~\cite{franco2023hyperbolic} & ICLR'23 & J & 78.2 & 82.6 & 61.8 & 64.6 &  \multicolumn{2}{c}{-} \\
			UmURL~\cite{sun2023unified} & ACM MM'23 & J & 82.3 & 89.8 & 73.5 & 74.3 &  \multicolumn{2}{c}{52.1} \\
			UmURL~\cite{sun2023unified} & ACM MM'23 & J+M+B  & 84.2  & 90.9  & 75.2  & 76.3  & \multicolumn{2}{c}{52.6} \\
			Heter-Skeleton~\cite{wang2025heterogeneous}  & CVPR'25 & J & 80.2 & 88.0 & 70.7 & 73.5 & \multicolumn{2}{c}{47.7} \\
			\textbf{USDRL (STTR)} & Preliminary work & J     & 84.2  & 90.8  & 76.0  & 76.9  & \multicolumn{2}{c}{51.8} \\
			\textbf{USDRL (DSTE)} & Preliminary work & J     & 85.2  & 91.7  & 76.6  & 78.1  & \multicolumn{2}{c}{54.4} \\
			\textbf{USDRL (STTR)} & This work & J+M+B & 85.8 & 91.8 & 77.5 &78.8 & \multicolumn{2}{c}{54.7} \\
			% \textbf{USDRL (DSTE)} & This work & J+M+B & \\
			\midrule
			\textit{\textbf{3s-ensemble}} &       &       &       &       &       &       & \multicolumn{2}{c}{} \\
			% 3s-HiCLR~\cite{zhang2023hierarchical} & AAAI'23 & J+M+B & 80.4  & 85.5  & 68.2  & 68.8  & \multicolumn{2}{c}{53.8} \\
			3s-CMD~\cite{mao2022cmd} & ECCV'22 & J+M+B & 84.1  & 90.9  & 74.7  & 76.1  & \multicolumn{2}{c}{52.6} \\
			3s-CPM~\cite{zhang2022contrastive}   & ECCV'22 & J+M+B     & 83.2  & 87.0  & 73.0  & 74.0  & \multicolumn{2}{c}{51.5}  \\
			3s-HiCLR~\cite{zhang2023hierarchical}   & AAAI'23 & J+M+B     & 80.4  & 85.5  & 70.0  & 70.4  & \multicolumn{2}{c}{-}  \\
			3s-SkeAttnCLR~\cite{hua2023part}   & IJCAI'23 & J+M+B     & 82.0  & 86.5  & 77.1  & 80.0  & \multicolumn{2}{c}{55.5}  \\
			3s-SSRL~\cite{jin2023ssrl}   & TCSVT'23 & J+M+B     & 81.6  & 85.1  & 69.2  & 71.5  & \multicolumn{2}{c}{50.2}  \\
			3s-PCM$^{\rm 3}$~\cite{zhang2023prompted}  & ACM MM'23 & J+M+B     & 87.4  & 93.1  & 80.0  & 81.2  & \multicolumn{2}{c}{58.2} \\
			3s-ActCLR~\cite{lin2023actionlet} & CVPR'23 & J+M+B  & 84.3 & 88.8 & 74.3 & 75.7 & \multicolumn{2}{c}{-} \\
			3s-RVTCLR+~\cite{zhu2023modeling} & ICCV'23  & J+M+B  & 79.7 & 84.6 & 68.0 & 68.9 & \multicolumn{2}{c}{-} \\
			3s-PSTL~\cite{zhou2023self}  & AAAI'23 & J+M+B     & 79.1  & 82.6  & 69.2  & 70.3  & \multicolumn{2}{c}{52.3} \\
			3s-CSTCN~\cite{wang2023learning}  & TMM'23 & J+M+B     & 85.8  & 92.0  & 77.5  & 78.5  & \multicolumn{2}{c}{53.9} \\
			3s-UmURL~\cite{sun2023unified} & ACM MM'23 & J+M+B & 84.4  & 91.4  & 75.8  & 77.2  & \multicolumn{2}{c}{54.3} \\
			% 3s-Skeleton-logoCLR~\cite{hu2024global}  & TCSVT'24 & J+M+B     & 86.1  & 89.8  & 79.8  & 80.1  & \multicolumn{2}{c}{57.7} \\
			% 3s-Eq-Contrast~\cite{lin2024mutual}  & TIP'24 & J+M+B     & 87.0  & 92.9  & 79.4  & 81.2  & \multicolumn{2}{c}{-} \\
			% \midrule
			% 
			\textbf{3s-USDRL (DSTE)} & This work & J+M+B & \textbf{87.1}   & \textbf{93.2}   & \textbf{79.3}   & \textbf{80.6}   & \multicolumn{2}{c}{\textbf{59.7}} \\
			\bottomrule
		\end{tabular}
		% }
	\label{linear}
\end{table*}
\subsection{Implementation Details}
\noindent{\textbf{Model Structure:}} 
For data process and augmentation, we adhere to the same strategies employed in works~\cite{wu2024scd, sun2023unified}. 
The DSTE Encoder is a two-layer architecture. The channel dimensions of embedding, representation, and projection are configured as 1024, 1024, and 2048, respectively. The projector consists of two linear layers, each followed by batch normalization and a ReLU activation, and a third linear layer. The output dimensions for the spatial, temporal, and instance projectors are 2048, 2048, and 4096, respectively.% Note that batch sizes of 256 and 512 yields similar performance.%, but larger batches reduce training time. %Additional experimental settings are provided in the supplementary materials. 

Our model comprises a skeleton encoder serving as the backbone, accompanied by three domain-specific projectors, each with a simple and identical structure consisting of several linear layers. We employ both STTR and DSTE as alternative backbones for our model, facilitating comparative analysis, each of these backbones consists of a two-layer architecture. For the NTU60/120 and PKU-MMD I datasets, the channel numbers of embedding, representation, and projection, $C_e$, $C_r$, and $C_p$, are set to 1024, 1024, and 2048, respectively. In contrast, for the PKU-MMD II dataset, which has fewer samples yet features longer, more complex action sequences than the NTU datasets, the dimensions are adjusted to 512, 512, and 1024 respectively.

\noindent{\textbf{Hyperparameters in Pre-training:}} During the pretraining stage, the $gap$ in the Dense Shift operation is set to 4. The hyperparameters for the weighted sum of the CA and DSA modules, $\alpha$ and $\beta$, are both set to 0.5. the hyperparameters $\tau$, used to balance the different domain terms in the pretrain loss $\mathcal{L}_{mfd}$, is also set to 0.5. In Intra-sample Consistency, the hyperparameters $\kappa$ and $\eta$ are set to 5 and 5e-4 respectively. In the case of Inter-sample separability, the hyperparameters $\mu$ and $\lambda$, used to balance the cross-correlation and variance terms, are set to 1 and 0.001 respectively.

\noindent{\textbf{Training Details:}}
For the optimizer, we employ the Adam with a weight decay of 1e-5. The batch size is set to 324. In practice, we observe that batch sizes of 256 and 512 yield similar performance metrics. However, larger batch sizes significantly reduce training time, suggesting a trade-off between computational efficiency and resource utilization. The model undergoes training for 450 epochs on the NTU60/120 datasets and 1200 epochs on the PKU-MMD II dataset. The initial learning rate is set to 5e-4 and reduced to 5e-5 at epoch 350 for NTU60/120 and epoch 1000 for PKU-MMD II, respectively. 

\noindent{\textbf{Baselines and Evaluations:}} 
% We also replace the proposed DSTE with Transformer-based STTR~\cite{plizzari2021skeleton} in our framework and report the performance.
We consider the transformer-based STTR~\cite{plizzari2021skeleton} with self-supervised training and plain feature decorrelation as the baseline. We gradually replace the training method with the proposed MG-FD and the backbone with the proposed DSTE.
% Finally, unified multimodal training is incorporated into the framework.

We evaluate the effectiveness of our method across eight downstream tasks, which are categorized into coarse prediction, dense prediction, and transfer prediction. Coarse prediction tasks include both unsupervised and semi-supervised action recognition, as well as action retrieval. The instance-level representation is obtained by applying a MaxPooling operation across frames and concatenating the spatial and temporal streams. Dense prediction tasks include action detection, segmentation and prediction, where frame-wise predictions are made using dense representations. Transferred prediction tasks involve transfer learning for both action recognition and retrieval. Since transfer learning for skeleton-based action retrieval has not been studied before, we set up a benchmark and implement several baselines to achieve this goal. 
% For action recognition and action retrieval, which are instance-level classification tasks, we employ MaxPooling and Concatenation to obtain the instance-level representations for evaluation. For action detection and segmentation, we perform frame-wise predictions using dense representations obtained by the backbone network.

\subsection{Results of Coarse Prediction Tasks}
Coarse prediction uses the learned global representation for action recognition and retrieval. We conduct experiments on three popular tasks: unsupervised action recognition, semi-supervised action recognition and skeleton-based action retrieval.

\noindent{\textbf{{Unsupervised Action Recognition:}} This task involves training an encoder for skeleton-based action recognition through unsupervised pre-training. After the pre-training phase, a fully connected layer is appended to the encoder's backbone network. During the evaluation process, the backbone network remains frozen, and only this newly added linear layer is fine-tuned to assess the quality of the representations learned during pre-training. 
Table \ref{linear} presents accuracies on the popular five benchmarks from NTU-60, NTU-120, and PKU-MMD II datasets. The comparative methods are classified into four groups according to different self-supervised learning methods: hybrid learning, masked sequence modeling, constrastive learning and feature decorrelation. 
The proposed USDRL achieves state-of-the-art performance across all types of comparative approaches, surpassing the previous state-of-the-art feature decorrelation-based method by an average margin of 2.8\% across all benchmarks. Specifically, USDRL utilizing only the joint modal consistently outperforms UmURL~\cite{sun2023unified} employing three modalities by approximately 1.4\%. In addition, the improved USDRL with multi-perspective consistency training outperforms the preliminary method by 1\% on the NUT-60 dataset and 2.6\% on the PKU-MMD II dataset. 
It can be concluded that, when compared with other self-supervised learning methods, such as \textit{Mask Sequence Modeling}, \textit{Negative-based Contrastive Learning}, and \textit{Hybrid Learning}, USDRL exhibits superior performance. Furthermore, training USDRL with additional bone and motion streams and subsequently conducting an ensemble of these models leads to a further substantial improvement.

\noindent{\textbf{{Semi-Supervised Action Recognition:}} In the semi-supervised setting, the pre-trained encoder is first loaded, and subsequently, the entire model is fine-tuned using only 1\% and 10\% of randomly sampled labeled training data. Results on the NTU-60 dataset are reported in Table \ref{semi}. 
% We compare our approach with existing works, specifically, LongT GAN~\cite{zheng2018unsupervised}, MS$^{2}$L~\cite{MS2L}, ISC~\cite{thoker2021skeleton}, HiCLR~\cite{hico2023}, CMD~\cite{mao2022cmd}, PCM$^{\rm 3}$~\cite{zhang2023prompted}. 

% With only 1\% of labeled training data, our method achieves accuracies of 57.3\% and 60.7\% on the x-sub and x-view protocols, respectively. It outperforms existing state-of-the-art methods by 4.3\% and 3.7\% for the 1\% data and 10\% data on the x-view protocol, respectively. These results confirm the strong generalization capability of our approach and demonstrate its competitive performance in semi-supervised action recognition.
In the scenario where a mere 1\% of the training data is labeled, our proposed method attains accuracies of 57.3\% and 60.7\% on the x-sub and x-view evaluation protocols, respectively. Specifically, when leveraging 1\% of labeled training samples, our method surpasses the current state-of-the-art approaches by a margin of 4.3\% on the x-view protocol. Moreover, when utilizing 10\% of labeled data for training, it further extends this performance advantage, outperforming the existing best methods by 3.7\% on the same x-view protocol. These results confirm the robust generalization capability of our approach and demonstrate its competitive performance in semi-supervised action recognition.

% Table generated by Excel2LaTeX from sheet 'linear'
\begin{table}[tb]
	\centering
		\caption{Comparison of performance under semi-supervised evaluation protocol on the NTU60 dataset.}
	\resizebox{0.49\textwidth}{!}{ 
		\begin{tabular}{lcccc}
			\toprule
			\multirow{2}[4]{*}{\textbf{Method}} & \multicolumn{2}{c}{\textbf{x-sub}} & \multicolumn{2}{c}{\textbf{x-view}} \\
			\cmidrule{2-5}          & 1~\% data & 10~\% data & 1~\% data & 10~\% data \\
			\midrule
			% LongT GAN~\cite{zheng2018unsupervised} & 35.2  & 62    & -     & - \\
			MS$^\text{2}$L~\cite{MS2L}  & 33.1  & 65.2  & -     & - \\
			ISC~\cite{thoker2021skeleton}   & 35.7  & 65.9  & 38.1  & 72.5 \\
			HiCLR~\cite{zhang2023hierarchical} & 51.1  & 74.6  & 50.9  & 79.6 \\
			CMD~\cite{mao2022cmd}   & 50.6  & 75.4  & 53    & 80.2 \\
			PCM$^{\rm 3}$~\cite{zhang2023prompted}  & 53.8  & 77.1  & 53.1  & 82.8 \\
			%SCD-Net & 69.1  & 82.2  & 66.8  & 85.8 \\
			% Skeleton-logoCLR~\cite{hu2024global}  & 68.8  & -  & 69.6  & - \\
			% 3s-Skeleton-logoCLR~\cite{hu2024global}  & -  & 82.4  & -  & 85.1 \\
			% 2s-DMMG~\cite{guan2023dmmg}  & 56.1  & 81.8  & 56.6  & 85.1 \\
			% CPM~\cite{zhang2022contrastive}  & 56.7  & 73.0  & 57.5  & 77.1 \\
			% CSTCN~\cite{wang2023learning}  & 75.7  & 81.0  & 71.6  & 83.7 \\
			% 3s-CSTCN~\cite{wang2023learning}  & 77.9  & 84.7  & 72.8  & 87.5 \\
			% 3s-Eq-Contrast~\cite{lin2024mutual}  & 60.8  & 81.7  & 60.7  & 86.0 \\
			% 3s-HiCLR~\cite{zhang2023hierarchical}  & 54.7  & 82.1  & 53.7  & 84.8 \\
			% 3s-SkeAttnCLR~\cite{hua2023part}  & 59.6  & 81.5  & 59.2  & 83.8 \\
			HiCo~\cite{hico2023}  & 54.4  & 73.0  & 54.8  & 78.3 \\
			% MacDiff~\cite{wu2024macdiff}  & 72.0  & 89.2  & 79.2  & 93.1 \\
			% MMFR~\cite{zhu2024motion}  & 65.0  & 87.0  & 71.3  & 91.1 \\
			% MAMP~\cite{mao2023masked}  & 66.0  & 88.0  & 68.7  & 91.5 \\
			Heter-Skeleton~\cite{wang2025heterogeneous} & 55.0 & 76.3 & 55.0 & 79.1 \\
			\midrule
			\textbf{USDRL (STTR)} & 55.0 & 76.1 & 59.1 & 82.0 \\
			\textbf{USDRL (DSTE)} & \textbf{57.3} & \textbf{80.2} & \textbf{60.7} & \textbf{84.0} \\
			\bottomrule
		\end{tabular}
	}
	\label{semi}
\end{table}
% Table generated by Excel2LaTeX from sheet 'linear'
\begin{table}[tbp]
  \centering
    \caption{\RED{Results of 2D skeleton-based action recognition on the UAV-Human dataset. S and U denote supervised and unsupervised training, respectively.}}
    \begin{tabular}{lcccc}
    \toprule
    \textbf{Method} & Modality & Training & CS-v1 & CS-v2 \\
    \midrule
    ST-GCN~\cite{yan2018spatial}	 & J & S & 30.2  & 56.1 \\
    2s-AGCN~\cite{shi2019two}	 & J+B & S & 34.8  & 66.7 \\
    HARD-Net~\cite{li2020hard}  & J & S & 37.0  & - \\
    Shift-GCN~\cite{cheng2020skeleton} & J & S & 38.0  & 67.0 \\
    LLM-AR~\cite{qu2024llms}  & J & S & 46.3 & - \\
    \midrule
    \textbf{USDRL (STTR)} & J & U & 31.7   & 50.2 \\
    \textbf{USDRL (DSTE)} & J & U & 36.3 & 60.8 \\
    \bottomrule
    \end{tabular}
  \label{uav_2d}
\end{table}

% Table generated by Excel2LaTeX from sheet 'linear'
\begin{table}[tbp]
	\centering
	\caption{Comparison of action retrieval results. * denotes the improved method of our USDRL.}
	\resizebox{0.45\textwidth}{!}{ 
		\begin{tabular}{lcccc}
			\toprule
			\multirow{2}[4]{*}{\textbf{Method}} & \multicolumn{2}{c}{\textbf{NTU-60}} & \multicolumn{2}{c}{\textbf{NTU-120}} \\
			\cmidrule{2-5}          & x-sub & x-view & x-sub & x-set \\
			\midrule
			% LongT GAN~\cite{zheng2018unsupervised} & 39.1  & 48.1  & 31.5  & 35.5 \\
			ISC~\cite{thoker2021skeleton}   & 62.5  & 82.6  & 50.6  & 52.3 \\
			HaLP~\cite{Shah_2023_CVPR}  & 65.8  & 83.6  & 55.8  & 59 \\
			CMD~\cite{mao2022cmd}   & 70.6  & 85.4  & 58.3  & 60.9 \\
			UmURL~\cite{sun2023unified} & 71.3  & 88.3  & 58.5  & 60.9 \\
			PCM$^{\rm 3}$~\cite{zhang2023prompted}  & 73.7  & 88.8  & 63.1  & \textbf{66.8} \\
			HiCo~\cite{hico2023}  & 68.3  & 84.8  & 56.6  & 59.1 \\
			Heter-Skeleton~\cite{wang2025heterogeneous} &  66.3 & 87.1 & 55.7 & 59.8 \\
			% SCD-Net~\cite{wu2024scd}  & 76.2  & 86.8  & 59.8  & 65.7 \\
			\midrule
			\textbf{USDRL (STTR)} & 73.7  & 88.5 & 58.9 & 64.8 \\
			\textbf{USDRL (DSTE)} & 75.0  & 89.3 & 63.3 & 66.7 \\
			\textbf{USDRL (STTR)*} & 74.4 & 93.5 & 62.4 & 65.6 \\
			\bottomrule
	\end{tabular}}
	\label{knn}
\end{table}

\noindent{\textbf{{\RED{2D Skeleton-Based Action Recognition:}}} \RED{In addition to 3D skeleton-based action recognition, we evaluate our approach for 2D skeleton-based action recognition. Results on the challenging UAV-Human~\cite{li2021uav} is shown in Table~\ref{uav_2d}. It can be observed that our method even outperforms some supervised training methods.}

\noindent{\textbf{{Skeleton-Based Action Retrieval:}} 
% For action retrieval, representations obtained from the pre-trained encoder are directly utilized for retrieval tasks without any additional training. Specifically, upon receiving an action query, the nearest neighbor is identified within the representation space using cosine similarity. Results on the NTU-60 and NTU-120 datasets are shown in Table \ref{knn}, which compare our method against various state-of-the-art approaches. Utilizing the joint modality for inference, our proposed model significantly outperforms previous works~\cite{zhang2023prompted, sun2023unified}, demonstrating the distinguishability and effectiveness of the features learned by our approach.
In this task, representations extracted from the pre-trained encoder are directly employed for retrieval tasks, obviating the need for any additional training. Specifically, the nearest neighbor is identified within the representation space by calculating the cosine similarity between the query representation and all the gallery representations. Results on the NTU-60 and NTU-120 datasets are presented in Table \ref{knn}, where our method is compared with various state-of-the-art approaches. By only leveraging the joint modality, our proposed model substantially outperforms previous works~\cite{zhang2023prompted, sun2023unified}, thereby underscoring the discernibility and efficacy of the features acquired through our approach.

% Table generated by Excel2LaTeX from sheet 'linear'
\begin{table}[tbp]
  \centering
    \caption{Comparsion of action detection results on PKU-MMD I xsub benchmark with an overlap ratio of 0.5.}
    \begin{tabular}{lcc}
    \toprule
    \textbf{Method} & mAP$_\text{a}$~(\%) & mAP$_\text{v}$~(\%) \\
    \midrule
    MS$^\text{2}$L~\cite{MS2L} & 50.9  & 49.1 \\
    CRRL~\cite{wang2022contrast}  & 52.8  & 50.5 \\
    ISC~\cite{thoker2021skeleton}   & 55.1  & 54.2 \\
    CMD~\cite{mao2022cmd}   & 59.4  & 59.2 \\
    PCM$^{\rm 3}$~\cite{zhang2023prompted}  & 61.8  & 61.3 \\
    \midrule
    \textbf{USDRL (STTR)} & 66.1 & 65.9 \\
    \textbf{USDRL (DSTE)} & \textbf{75.7} & \textbf{74.9} \\
    \bottomrule
    \end{tabular}
  \label{detection}
\end{table}

\begin{table}[tbp]
  \centering
    \caption{Comparsion of action prediction results on the NTU-60 dataset.}
    \resizebox{0.5\textwidth}{!}{ 
    \begin{tabular}{lcccccccccc}
    \toprule
    \textbf{Method} &  10\% & 20\% & 30\% & 40\% & 50\% & 60\% & 70\%  & 80\% & 90\% & 100\% \\
    \midrule
    DeepSCN~\cite{kong2017deep} & 16.8 & 21.5 & 30.5 & 39.9 & 48.7 & 54.6 & 58.2 & 60.2 & 60.0 & 58.6 \\
    MSRNN~\cite{hu2018early} & 15.2 & 20.3 & 29.5 & 41.4 & 51.6 & 59.2 & 63.9 & 67.4 & 68.9 & 69.2 \\
    P-TSL~\cite{wang2019progressive} & 27.8 & 35.8 & 46.3 & 58.5 & 67.4 & 73.9 & 77.6 & 80.1 & 81.5 & 82.0 \\
     \midrule
    \textbf{USDRL (STTR)} &  24.7 & 36.5 & 54 & 65.7 & 72.8	& 76.9 & 80.3 & 82.4 & 83.7 & 84.2 \\
    \textbf{USDRL (DSTE)} &  25.5  & 36.8 & 54.8 & 66.7	 & 73.6	& 77.9	& 81.4	& 83.6 & 84.5	& 85.2 \\ 
    \bottomrule
    \end{tabular}}
  \label{prediction}
\end{table}

\begin{table}[tbp]
	\centering
	\caption{Comparison of action segmentation results on the PKU-MMD II dataset.}
    \begin{tabular}{l|ccccc}
     \toprule 
	Method & Acc & Edit & F10 & F25 & F50 \\
	 \midrule
	ST-GCN~\cite{yan2018spatial} & 64.9 & - & - & - & 15.5 \\
	MS-TCN~\cite{farha2019ms} & 65.5 & - & - & - & 46.3 \\
	ETSN~\cite{li2021efficient} & 68.4 & 67.1 & 70.4 & 65.5 & 52.0 \\
	% ASRF & 67.7 & 67.1 & 72.1 & 68.3 & 56.8 \\
	% MS-GCN & 68.5 & - & - & - & 51.6 \\
	CTC~\cite{xu2023efficient} & 69.2 & - & 69.9 & 66.4 & 53.8 \\
	DeST~\cite{li2023decoupled} & 67.6 & 66.3 & 71.7 & 68 & 55.5 \\
	% LaSA & ECCV2024 & 73.5 & 73.4 & 78.3 & 74.8 & 63.6 \\
     \midrule
	\textbf{USDRL (STTR)} &  68.7 & 67.5 & 70.9 & 67.3 & 56.2 \\
    \bottomrule
\end{tabular}
\label{segmentation}
\end{table}

\subsection{Results of Dense Prediction Tasks}
Dense prediction aims to learn frame-wise representation for skeleton-based action understanding. We perform experiments on three important tasks: action detection, action segmentation and action prediction. 

\noindent{\textbf{{Skeleton-Based Action Detection:}} We evaluate the performance of our model for action detection on the PKU-MMD I dataset. Following the standard protocols delineated in~\cite{chen2022hierarchically, liu2020benchmark}, we adopt two metrics: the mean average precision (mAP) for different actions (denoted as mAP${_\text{a}}$) and the mAP for different videos (denoted as mAP${_\text{v}}$). These metrics are computed with a default overlap ratio threshold of 0.5. As presented in Table~\ref{detection}, our approach achieves a remarkable performance improvement, significantly surpassing the state-of-the-art methods on this dataset. For example, when employing the same backbone of STTR, our approach outperforms CMD~\cite{mao2022cmd} and PCM$^{\rm 3}$~\cite{zhang2023prompted} by 6.7\% and 4.3\% in terms of mAP${_\text{a}}$, respectively. Furthermore, our approach with the proposed DSTE backbone also surpasses the variant using STTR by nearly 9\%. These results accentuate the model's adeptness in acquiring dense and discriminative feature representations, which are pivotal for precise action detection.
% Building upon the protocols established in~\cite{chen2022hierarchically, liu2020benchmark}, we evaluate the performance of our model for action detection on the PKU-MMD I dataset. Our approach involves attaching a linear classifier to the pre-trained encoder and fine-tuning the entire model to perform frame-wise action recognition. This encoder was initially pre-trained on the NTU 60 x-sub dataset and subsequently adapted to the PKU-MMD I dataset. We utilized the mean average precision (mAP) of different actions (mAP${_\text{a}}$) and different videos (mAP${_\text{v}}$), with an overlap ratio of 0.5, as evaluation metrics. As shown in Table~\ref{detection}, our approach significantly surpasses state-of-the-art methods and also outperforms the variant using STTR by nearly 9\%. These results demonstrate the effectiveness of the model in learning dense representations. 
%Detailed comparisons of action detection performance are in the supplemental materials.
% our proposed DSTE achieves an mAP${_\text{a}}$ of 75.7 \% and an mAP${_\text{v}}$ of 74.9 \%. Compared to the Spatial-Tmeporal Transformer network (STTR) used in~\cite{sun2023unified, wu2024scd}, DSTE outperforms STTR by nearly 9\% and also surpasses other state-of-the-art methods. We provide a detailed comparison of the detection performance between STTR and DSTE in the supplemental materials.

\noindent{\textbf{{Skeleton-Based Action Prediction:}} For action prediction, we report the action recognition accuracies with the observation ratio ranging from 10\% to 100\% in Table~\ref{prediction}. The state-of-the-art P-TSL~\cite{wang2019progressive} trains 10 distinct offline action recognition models, each corresponding to one of 10 different observation ratios. In contrast, our causal model employs a single unified architecture for online frame-wise action recognition, offering superior scalability and efficiency compared to existing offline models. The predicted probabilities of observed frames are simply aggregated for action recognition. Although our approach initially underperforms the offline state-of-the-art P-TSL~\cite{wang2019progressive} at an observation ratio of 10\%, it surpasses P-TSL’s performance when the observation ratio reaches 20\% or higher. Specifically, it outperforms P-TSL~\cite{wang2019progressive} by 6.2\% absolute accuracy when the observation ratio is 50\%, which demonstrates the potential of our model for online action understanding.

\noindent{\textbf{{Skeleton-Based Action Segmentation:}} For action segmentation, we evaluate performance using the following metrics: frame-wise accuracy (Acc), segmental edit score (Edit), and segmental F1 scores computed at intersection-over-union thresholds of 0.10, 0.25, and 0.5 (denoted as F1@10, F1@25, and F1@50, respectively). Results on the cross-subject evaluation of the PKU-MMD  II dataset are summarized in Table~\ref{segmentation}. Our approach outperforms the DeST~\cite{li2023decoupled} by 1.1\% and 1.2\% for the Acc and Edit metrics, respectively. Although our approach is not specifically designed for skeleton-based action segmentation, its performance on this task is quite competitive. 

% Table generated by Excel2LaTeX from sheet 'linear'
\begin{table}[tbp]
  \centering
   \caption{Comparisons of transferred action recognition results on the PKU-MMD II.}
    \begin{tabular}{lcc}
    \toprule
    \multirow{2}[4]{*}{\textbf{Method}} & \multicolumn{2}{c}{\textbf{Transfer to PKU-MMD II}} \\
\cmidrule{2-3}          & \textbf{NTU-60} & \textbf{NTU-120} \\
    \midrule
    % LongT GAN~\cite{zheng2018unsupervised} & 44.8  & - \\
    MS$^\text{2}$L~\cite{MS2L}  & 45.8  & - \\
    ISC~\cite{thoker2021skeleton}   & 45.9  & - \\
    SCD-Net~\cite{wu2024scd} & 56.3  & - \\
    HiCo~\cite{hico2023}  & 56.3  & 55.4 \\
    CMD~\cite{mao2022cmd}    & 56.0    & 57.0 \\
    UmURL~\cite{sun2023unified} & 58.2  & 57.6 \\
    \midrule
    \textbf{USDRL} & 57.2 & \textbf{58.3} \\
    \bottomrule
    \end{tabular}
  \label{ntu2pku}
\end{table}

\begin{table}[tbp]
	\centering
	\caption{Comparisons of transferred action retrieval results on the PKU-MMD II.}
	\begin{tabular}{lcccc}
		\toprule
		\multirow{3}[4]{*}{\textbf{Method}} &  \multicolumn{4}{c}{\textbf{Transfer to PKU-MMD II}} \\
		\cmidrule{2-5}
		& \multicolumn{2}{c}{\textbf{NTU-60}} & \multicolumn{2}{c}{\textbf{NTU-120}} \\
		\cmidrule{2-5}      & x-sub & x-view & x-sub & x-set \\
		\midrule
		UmURL~\cite{sun2023unified} & 43.4  & 42.8 & 42.5 & 43.0\\
		\textbf{USDRL} & 44.4 & 44.7	& 44.0 & 43.8 \\
		\bottomrule
	\end{tabular}
	\label{ntu2pku2}
\end{table}

\subsection{Results of Transferred Prediction Tasks}
Transferred prediction aims to assess the model's capability to generalize across different datasets for the purpose of action understanding. We evaluate our approach on two tasks: transfer learning for action recognition and transfer learning for action retrieval.

\noindent{\textbf{Transfer Learning for Action Recognition:}}
% We evaluate the generalizability of the learned representation by transferring knowledge from a source dataset to a target dataset. Specifically, we adopt the same settings as previous methods~\cite{MS2L, mao2022cmd, sun2023unified}, using NTU-60 and NTU-120 as the source datasets and PKU-MMD II as the target dataset. The evaluation follows the x-sub protocol. As presented in Table \ref{ntu2pku}, the proposed method achieves accuracies of 57.2\% when transferred from NTU-60 and 58.3\% from NTU-120, respectively, outperforming other methods. These results demonstrate that our framework effectively learns skeleton representations that generalize well to new datasets, which is crucial for real-world applications.
We assess the generalizability of the learned representations by transferring knowledge from a source dataset to a target one. Specifically, we adhere to the same experimental settings as previous studies~\cite{MS2L, mao2022cmd, sun2023unified}, employing NTU-60 and NTU-120 as source datasets and PKU-MMD II as the target dataset. The evaluation is conducted following the x-sub protocol. As shown in Table \ref{ntu2pku}, our proposed method attains an accuracy of 57.2\% when transferring knowledge from NTU-60 and 58.3\% when transferring from NTU-120, surpassing other competing methods. These results indicate that our framework effectively learns skeleton representations that generalize well to unseen datasets, a quality that is vital for real-world applications.

\begin{table}[tbp]
  \centering
   \caption{Ablation studies on the \textit{VAC}, \textit{XC}, \textit{MG-FD}, and the DSTE encoder under the x-sub evaluation on the NTU-60.}
  % The performance is evaluated on the NTU-60 X-sub under the linear evaluation protocol.
  % \resizebox{0.45\textwidth}{!}{ 
    \begin{tabular}{cccc|cc}
    \toprule
    \textit{VAC}   & \textit{XC} & \textit{MG-FD} & Encoder & Recog. & Retrieval \\
    \midrule
    \XSolidBrush     & \XSolidBrush  & \XSolidBrush    & STTR & 74.8   & 63.8 \\
    \Checkmark     & \XSolidBrush  & \Checkmark   & STTR & 83.7  & 73.2 \\
    \Checkmark     & \Checkmark   & \Checkmark  & STTR & \textbf{84.2}   & \textbf{73.7} \\
    \midrule
    \XSolidBrush     & \XSolidBrush   & \XSolidBrush   & DSTE & 77.3   & 64.4 \\
    \Checkmark     & \XSolidBrush   & \Checkmark  & DSTE  & 84.8  & 74.5 \\
    \Checkmark     & \Checkmark   & \XSolidBrush  & DSTE  & 84.6 & 74.7 \\
    \Checkmark     & \Checkmark   & \Checkmark  & DSTE  & \textbf{85.2} & \textbf{75.0} \\
    \bottomrule
    \end{tabular}
% }
  \label{tab:abl_loss_encoder}
\end{table}

\noindent{\textbf{Transfer Learning for Action Retrieval:}} We setup the benchmark of transfer learning for action retrieval. We use the self-supervised models trained on the NTU-60 and NTU-120 datasets directly for action retrieval on the PKU-MMD II dataset on Table~\ref{ntu2pku2}. Our model that is trained with the single joint modality consistently performs better than the multimodal model that is trained with joint, bone and motion modalities. These experimental results further demonstrate the generalization ability of our model.

\begin{figure*}[t]
	\centering
	\subfloat[Negative CL\label{subfig:negCL}]{
		\includegraphics[scale=0.225]{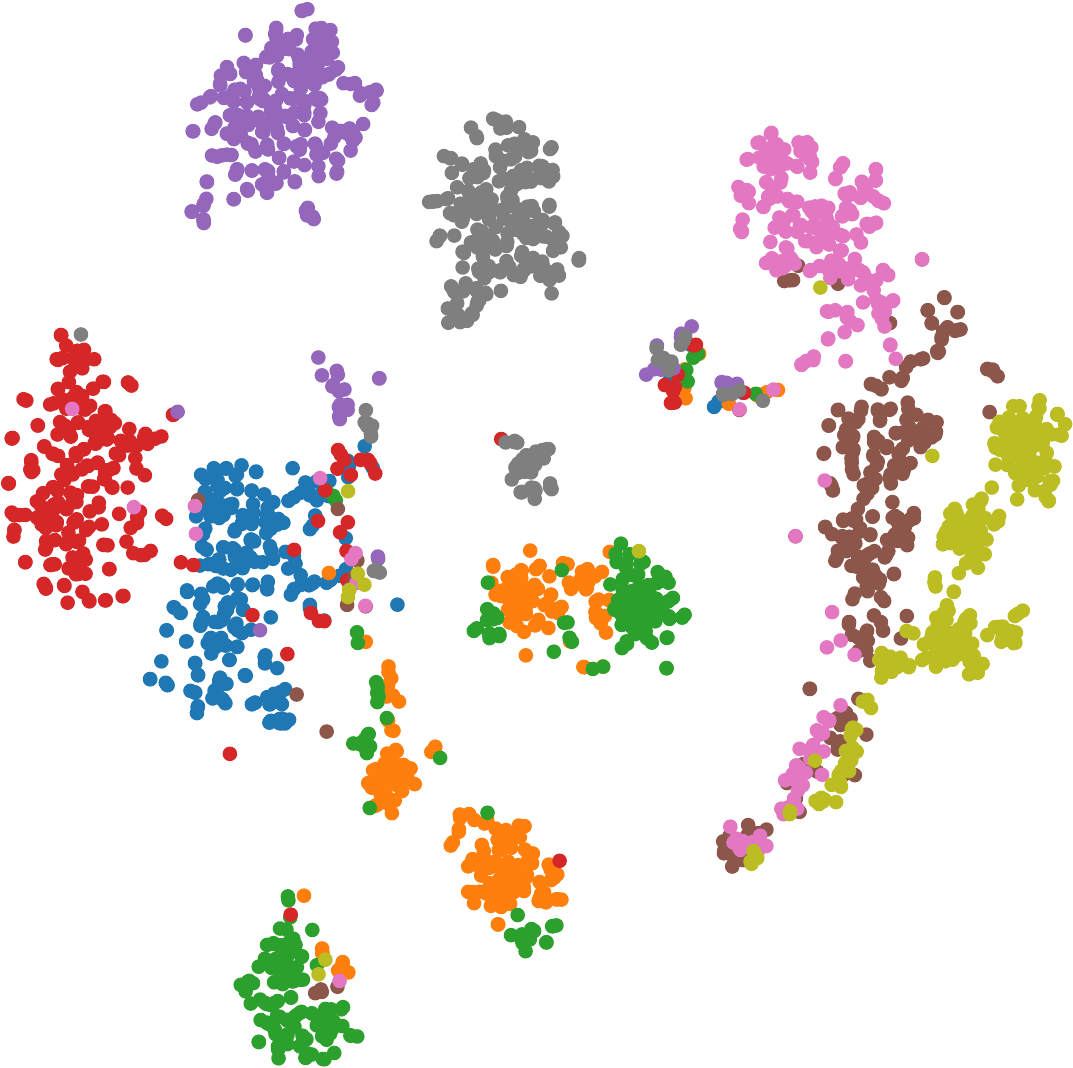}}
	\hspace{0.2cm}
	\subfloat[MG-FD w/o $XC$\label{subfig:woxc}]{
		\includegraphics[scale=0.225]{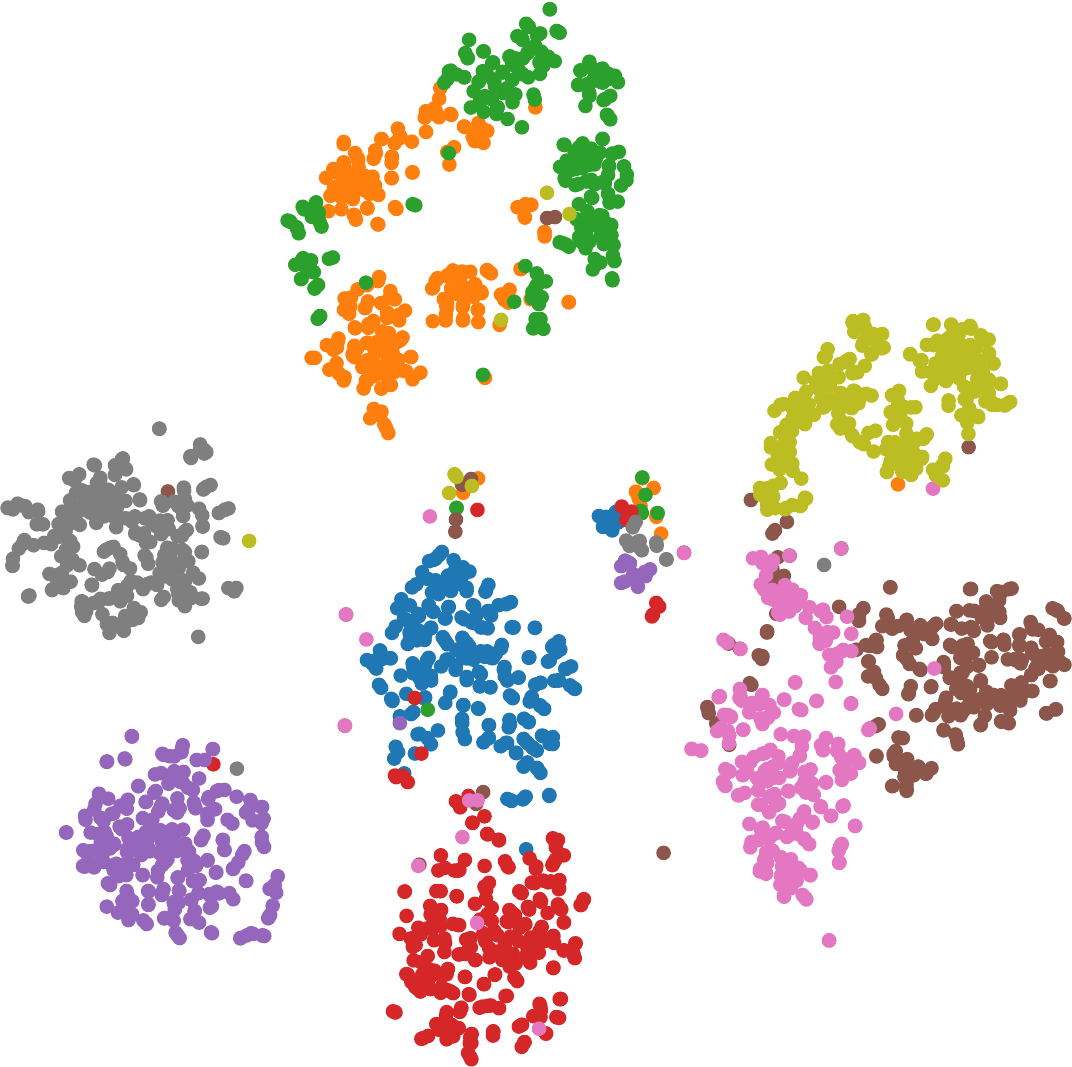}}
	\hspace{0.2cm}
	\subfloat[FD w/ $XC$\label{subfig:wost}]{
		\includegraphics[scale=0.225]{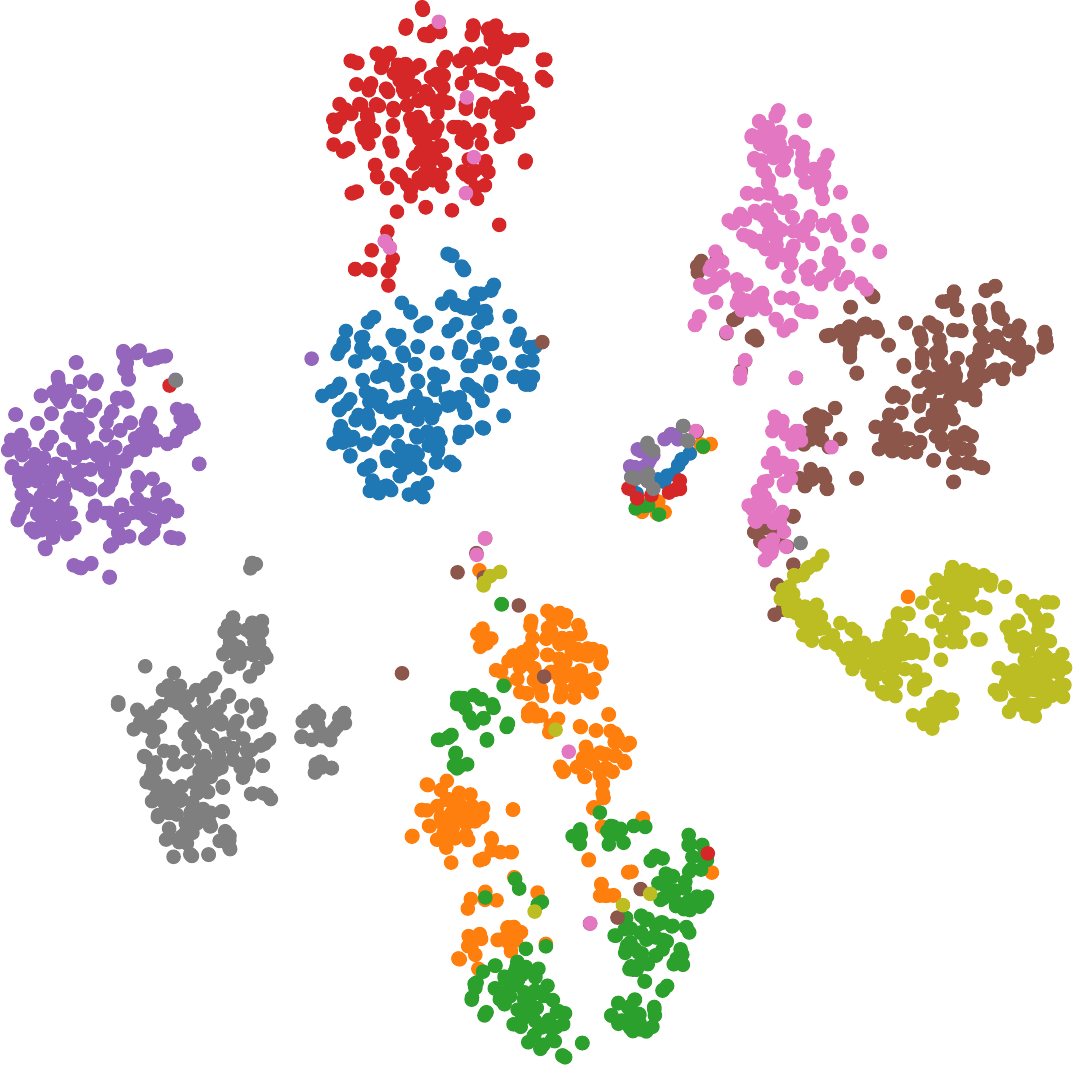}} % Single-Grained 
	\hspace{0.2cm}
	\subfloat[MG-FD\label{subfig:wst}]{  % w/ $XC$
		\includegraphics[scale=0.225]{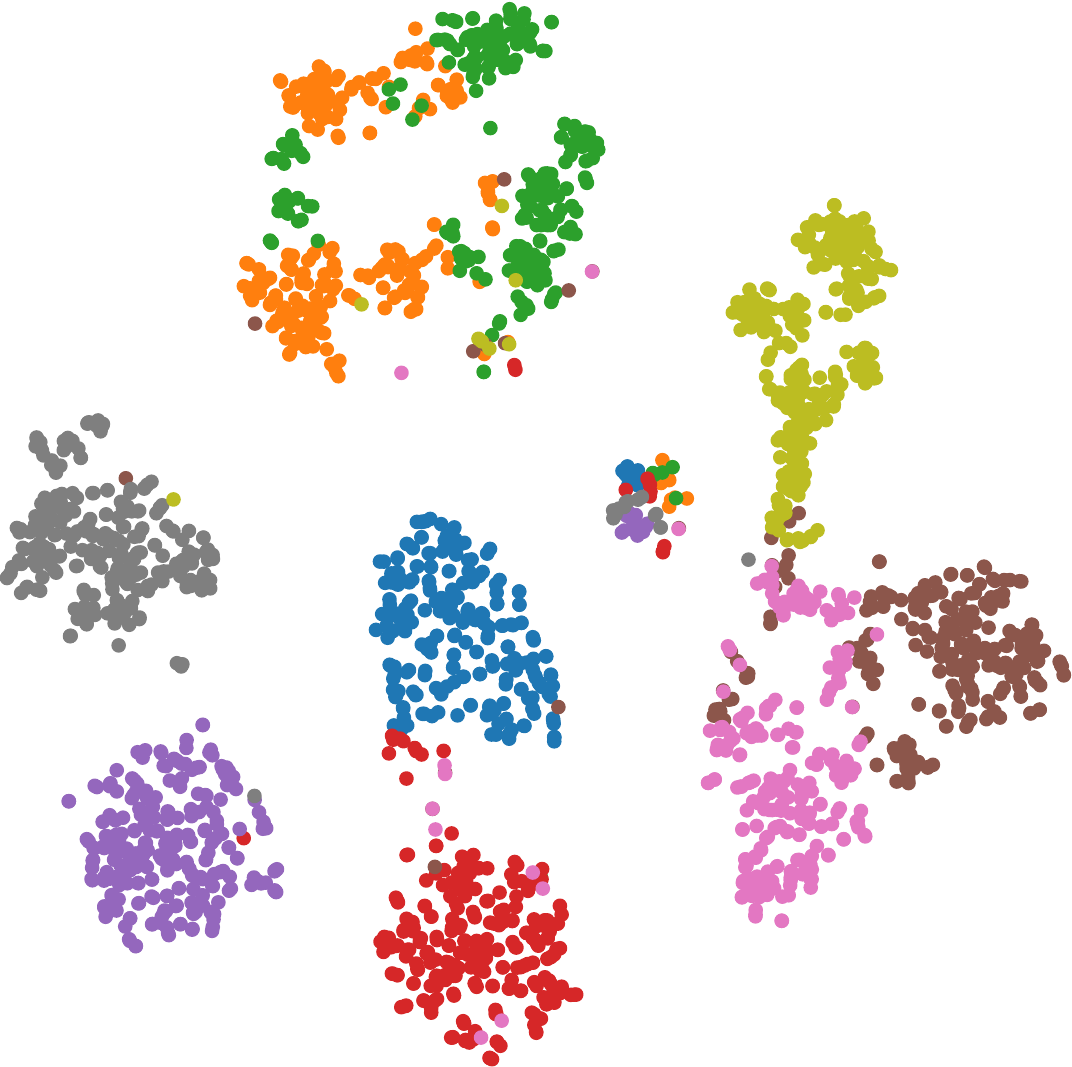}}
	\caption{Visualizations of learned instance-level representations obtained by (a) Contrastive Learning (CL), (b) Multi-Grained Feature Decorrelation (MG-FD) w/o $XC$, (c) Single-Grained FD w/ $XC$, and (d) MG-FD on the NTU-60. Nine classes from the testing set are randomly selected, and dots of the same color represent actions of the same class. We find that the model's ability to learn features that distinguish between different classes declines if multi-grained modeling or the cross-correlation ($XC$) term are omitted.
	}
	\label{fig:tsne}
\end{figure*}

\begin{figure}
	\centering
	\includegraphics[width=.95\linewidth]{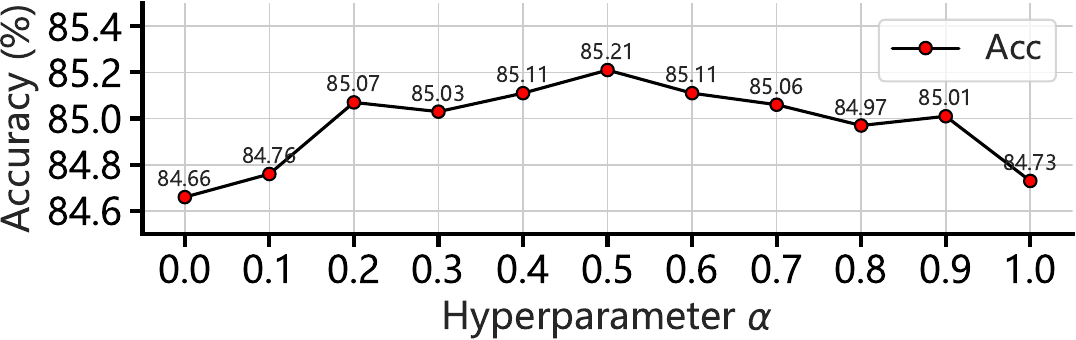}
	\caption{
		The impact of weight hyperparameter $\alpha$ for action recognition on the xsub evaluation of the NTU-60 dataset.
	} 
  \label{fig:alpha_beta}
\end{figure}

\begin{figure}[h]
	\centering
	\includegraphics[width=.98\linewidth]{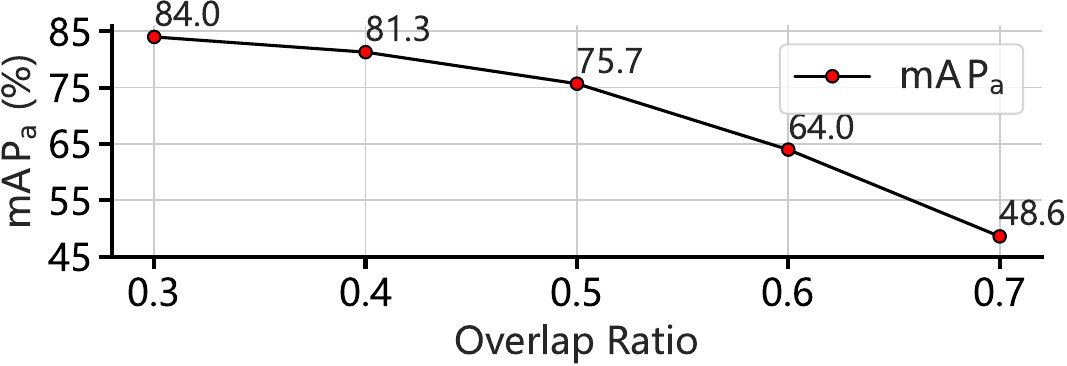}
	\caption{
		Mean Average Precision (mAP$\rm _a$) under different overlap ratios for action detection, evaluated on the PKU-MMD I dataset.
	} 
	\label{fig: map}
\end{figure}

\begin{figure*}
	\subfloat[]{\includegraphics[width=.95\linewidth]{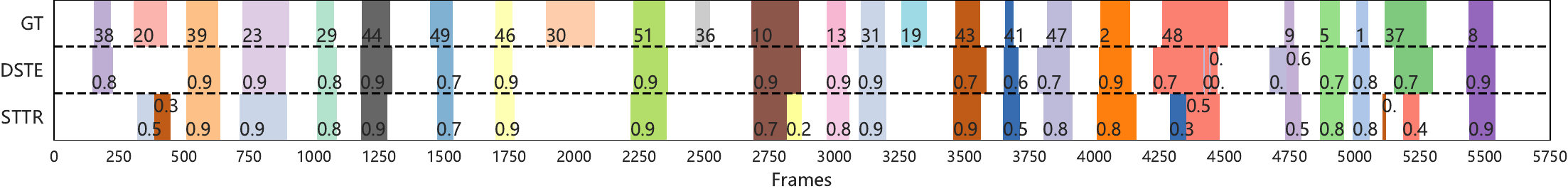}}\vspace{0pt}
	\subfloat[]{\includegraphics[width=.95\linewidth]{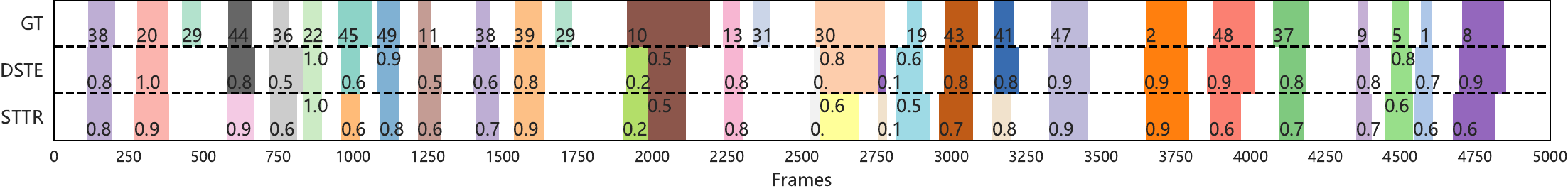}}\vspace{0pt}
	\subfloat[]{\includegraphics[width=.95\linewidth]{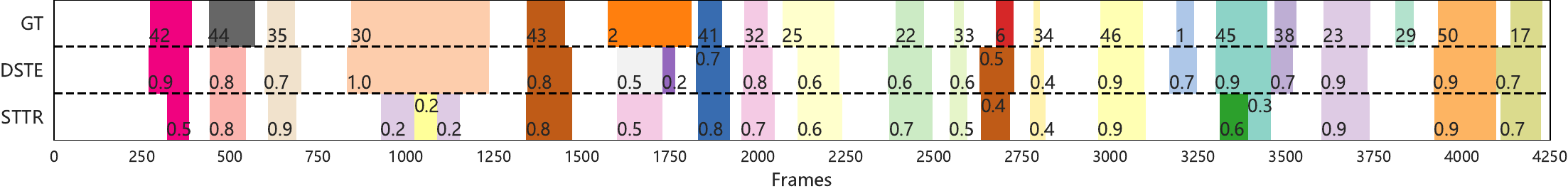}}\vspace{0pt}
	\subfloat[]{\includegraphics[width=.95\linewidth]{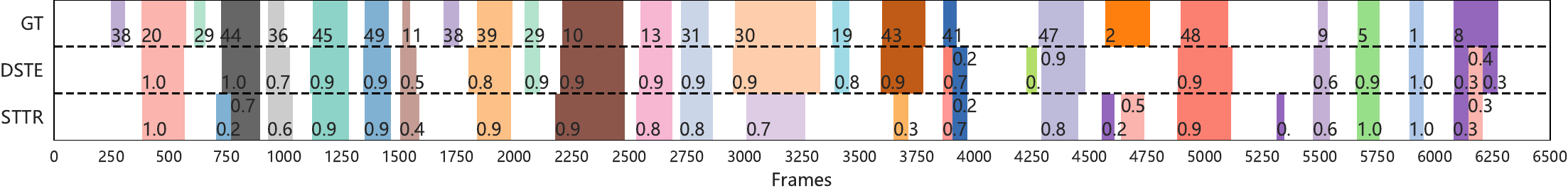}}\vspace{0pt}
	\subfloat[]{\includegraphics[width=.95\linewidth]{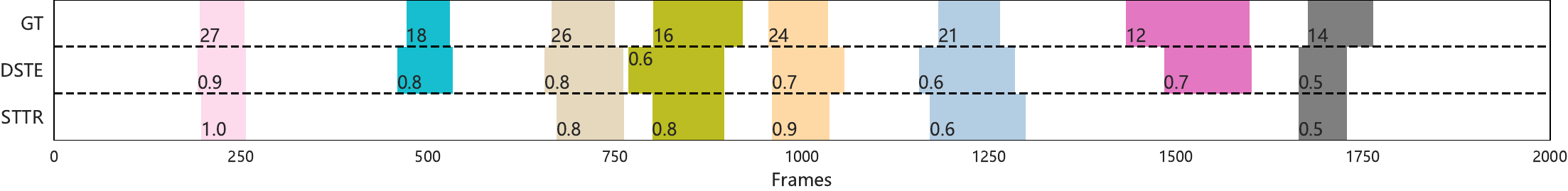}}\vspace{0pt}
	\caption{
		Visualization of action detection in video sequences on the PKUMMD dataset. We provide results of our approach utilizing both the proposed DSTE backbone and the STTR~\cite{plizzari2021skeleton} backbone. The number on the ground-truth segments denotes the index of the corresponding action class, whereas the numbers on the predicted segments represent the Intersection over Union (IoU) scores, where higher scores imply more accurate detections. Each sequence is visualized by colored segments, with each unique color distinguishing a distinct action.
	} 
	\label{fig: action_detection_vis}
\end{figure*}
% scale=0.18
\begin{figure*}
	\centering
	\subfloat[Top 20 Easy Categories]{
		\includegraphics[width=.5\linewidth]{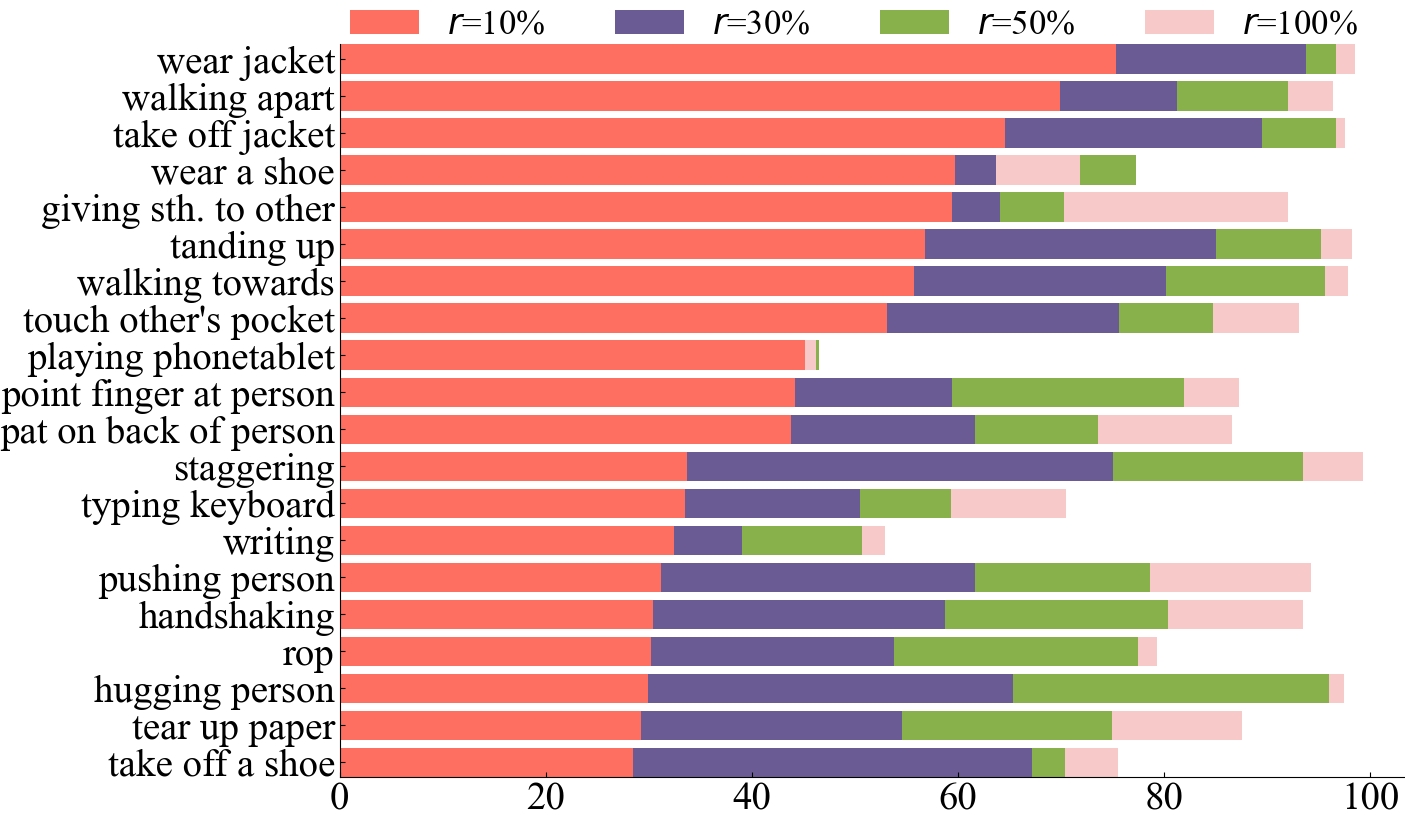}}
	\subfloat[Top 20 Difficult Categories]{
		\includegraphics[width=.5\linewidth]{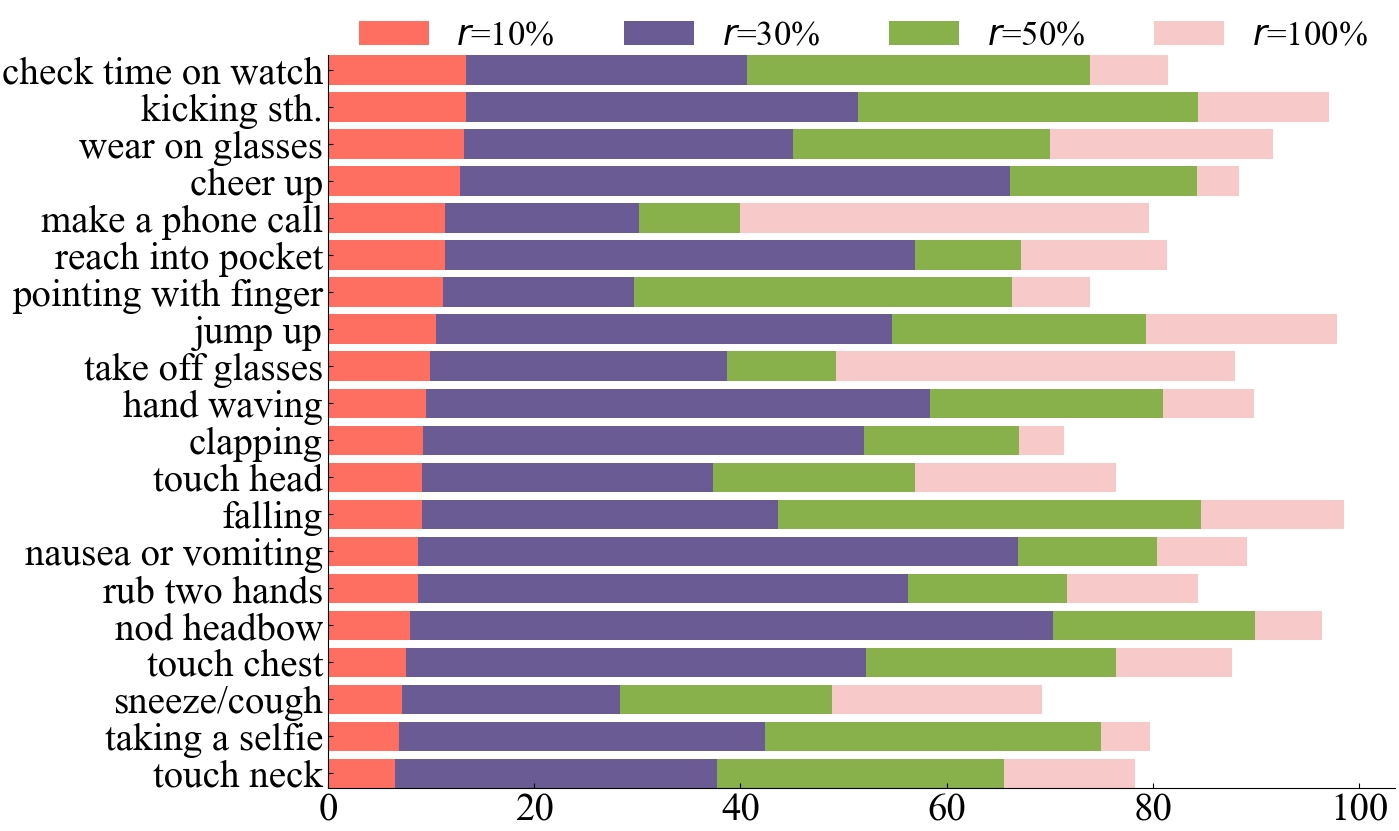}}
	\caption{
		Analysis of recognition accuracy for different actions in skeleton-based action prediction on the NTU-60 dataset. $r$ denotes the observation ratio, while the horizontal axis represents the recognition accuracy. We select the top easy and difficult action categories in (a) and (b), respectively. \RED{As the observation ratio increases, the accuracy of action recognition improves. With only the first 30\% of the observation, our method can accurately recognize most actions; even for the hard categories, the accuracy reaches 40\% (while random guessing accuracy is 1.6\%). }
		}
		% This demonstrates the potential of our method for early action recognition in applications like condition monitoring and human–computer interaction.
	\label{fig:cat_prediction} 
\end{figure*}

\subsection{Ablations, Parameter Analysis and Visualizations}
\noindent{\textbf{{Effectiveness of the VAC, XC and MG-FD:}} 
Multi-Grained Feature Decorrelation (MG-FD) consists of Variance and Auto-Covariance (VAC) term  and Cross-Correlation (XC) term, and each term is regularized on multi-grained representations (MG-FD). Table~\ref{tab:abl_loss_encoder} presents detailed ablation studies on the proposed method. Without VAC, XC and MG-FD denotes the baseline using traditional negative-based pretraining. We find that the results obtained using traditional negative-based pretraining strategies are significantly lower than those achieved with the feature decorrelation method. This demonstrates the strong potential and superiority of the feature decorrelation. 
Additionally, for both recognition and retrieval tasks, the method with \textit{XC} matrix improves 0.5\% over that without \textit{XC} matrix. 
This indicates that the \textit{XC} matrix could effectively capture and reduce the feature correlation between different augmented samples and help to learn more effective and robust representations. 
Combined with MG-FD, our approach further enhances the model's generalization ability. This improvement is attributed to the utilization of fine-grained features in the temporal, spatial and instance domains, as opposed to relying solely on the instance domain.

To further analyze the effectiveness of the proposed self-supervised representation learning paradigm, we visualize and compare learned features of different methods. As shown in Figure \ref{fig:tsne}, visualizations demonstrates distinct clustering of learned features of the proposed MG-FD, suggesting that they are capable of effectively separating different classes. These results imply that the learned features capture the intrinsic characteristics of each class, enabling clear discrimination among various action categories. 
% our method is capable of learning more discriminative features among different classes. 
% Combined with the visualizations in Figure \ref{fig: tsne}, our Multi-Grained Feature Decorrelation (MG-FD), by decorrelating across three domains, can further enhance accuracy. 
% For a more comprehensive ablation study, we also conduct additional experiments with our DSTE and STTR. Our DSTE outperforms the STTR under the same pretraining strategies, demonstrating its strong capability in modeling skeleton sequences.

%the Spatial-Temporal Transformer network (STTR) used in~\cite{sun2023unified,wu2024scd}. Our DSTE outperforms the STTR under the same pretraining strategies, demonstrating its strong capability in modeling skeleton sequences.

% To evaluate the effectiveness of the CA and DSA modules, 
% We conduct several ablation studies on the DSTE structure, focusing on the number of DSTE layers. 
\noindent{\textbf{{Effectiveness of the CA and DSA:}} The proposed DSTE backbone consists of the DSA and CA modules. We evaluate the influence of the weight hyperparameters $\alpha$ in Eq.~\ref{DSA} in the DSTE structure. 
As shown in Figure \ref{fig:alpha_beta}, $\alpha$ takes values in the range [0, 1] with an interval of 0.1, where 0 and 1 represent the isolated modules of DSA and CA, respectively. We observe that both DSA and CA produce significantly inferior results compared to other settings. These results underscore the efficacy and complementarity of CA and DSA modules.

% Table generated by Excel2LaTeX from sheet 'linear'
\begin{table}[tbp]
  \centering
   \caption{The study on the impact of the number of encoder layers. The performance is evaluated on the NTU60 X-sub dataset under the linear evaluation protocol.}
    \begin{tabular}{ccccc}
    \toprule
    Layer Num & Para. (M) & Recognition & Retrieval \\
    \midrule
    1     & 67.7 & 84.5     & 73.1 \\
    2     & 94.9 & 85.2  & 75.0 \\
    3     & 113.9 & 84.9      & 75.4 \\
    \bottomrule
    \end{tabular}
  \label{tab:abl_encoder_layer}
\end{table}

\noindent{\textbf{Impact of the Number of Encoder Layers:}}
We investigate the impact of the number of encoder layers on the performance in Table \ref{tab:abl_encoder_layer}. As the number of layers increases, the complexity of the model increases linearly. Experimental results suggest that a configuration with two layers of DSTE achieves an optimal balance between performance and complexity. 

\begin{table}[tbp]
  \centering
   \caption{Comparison of computation costs among state-of-the-art skeleton-based action recognition methods. The performance is evaluated on the NTU60 X-sub dataset.}
    \begin{tabular}{lccc}
    \toprule
    Method & Modality &  FLOPs/G & Recognition \\
    \midrule
    3s-PSTL~\cite{zhou2023self} & J+M+B  & 3.45 & 79.1 \\
    3s-CrosSCLR~\cite{li20213d} & J+M+B & 17.28 & 82.1 \\
    3s-CMD~\cite{mao2022cmd} &  J+M+B & 17.28 & 84.1 \\
    % UmURL~\cite{sun2023unified}  & J & 1.74 & 82.3 \\
    UmURL~\cite{sun2023unified}  & J+M+B & 2.54 & 84.2 \\
    \midrule
    USDRL (STTR)  & J &  1.74 & 84.2 \\
    USDRL (STTR)  & J+M+B & 2.54 & 85.8 \\
    \bottomrule
    \end{tabular}
  \label{tab:comp_cost}
\end{table}

\noindent{\textbf{{Comparison of Computation Costs:}} Table \ref{tab:comp_cost} compares the computational complexity and recognition accuracy of state-of-the-art skeleton-based action recognition models. It can be seen that our model can achieve the highest recognition accuracy with a very small amount of computational cost. For example, our model using single joint modality slightly outperforms the 3s-CMD~\cite{mao2022cmd}, while the computational cost of ours is only 10\% of that of this method.
	
\noindent{\textbf{Results of Action Detection with Different IoUs:}}
We evaluate the performance across different overlap ratios, ranging from 0.3 to 0.7, as shown in Figure \ref{fig: map}. Our approach achieves an mAP$_a$ of 48.6\% at an overlap ratio of 0.3 and 75.7\% at an overlap ratio of 0.5. These results demonstrate the efficacy of our method in dense prediction tasks by capturing fine-grained features. The robust performance across various overlap thresholds highlights our approach's adaptability, particularly in challenging scenarios. By excelling in both lower and higher overlap ratios, our model proves to be versatile, effectively adapting to different levels of temporal and spatial granularity required for accurate action detection in complex video sequences.

\noindent{\textbf{Visualization of Action Detection:}}
To further demonstrate the superior performance of our approach, along with the proposed backbone DSTE, over STTR~\cite{plizzari2021skeleton} in dense prediction tasks, we select a subset of video results, visualizing the action detection as depicted in Figure \ref{fig: action_detection_vis}. Our approach with the DSTE not only detects a greater number of action instances but also achieves more precise localization, as indicated by the higher Intersection over Union (IoU) scores.
% for both short and long action sequences.

\noindent{\textbf{Analysis of Performance for Action Prediction:}} Given that action prediction serves as a representative dense prediction task with significant real-world applications, we analyze and compare the accuracies across different actions in Figure \ref{fig:cat_prediction}. We observe that certain actions, such as wear jacket, walking apart, and take off jacket, can be accurately predicted at an early stage, specifically within the first 10\% of the observed sequence. When 50\% of the sequence is observed, most actions can be recognized early with high accuracy, and even the more challenging actions achieve accuracies exceeding 60\%, with only a few exceptions. This demonstrates the potential of our method for early action recognition in applications like condition monitoring and human–robot interaction.
% These findings indicate the potential of the proposed method for action prediction in real-world scenarios.

\section{Conclusion}
% We introduce a novel Unified Skeleton-based Dense Representation Learning (USDRL) framework that leverages feature decorrelation to address skeleton-based downstream tasks. Unlike most existing contrastive learning approaches that rely on negative samples, our proposed USDRL eliminates the need for an additional momentum encoder and memory bank, significantly streamlining the complex pipeline associated with skeleton-based representation learning. Furthermore, our Dense Spaito-Temporal Encoder enhances the capability of our method to perform dense prediction tasks such as action detection effectively. We believe that USDRL offers a streamlined and effective alternative to negative-based contrastive learning methods in self-supervised skeleton-based representation learning.
We introduce a transformer-based foundation model for skeleton-based human action understanding. The proposed model comprises a Dense Spatio-Temporal Encoder (DSTE), Multi-Grained Feature Decorrelation (MG-FD), and Multi-Perspective Consistency Training (MPCT). We categorize skeleton-based human action understanding tasks into three types: coarse prediction, dense prediction, and transferred prediction, noting that dense prediction tasks have been largely overlooked. We conduct extensive experiments across eight tasks spanning the three categories and demonstrate that the proposed method serves as a unified and strong baseline for addressing them. Detailed ablation studies and analyses demonstrate the effectiveness of each module, highlighting the proposed method's high performance, strong generalization capability, and low computational cost. We believe this work could broaden the scope of research in skeleton-based action understanding and facilitate progress in dense prediction tasks, including action detection, action segmentation, and action prediction. \RED{A limitation of our approach is the need to fine-tune the linear classifier when adapting to new action understanding tasks, which limits its applicability in open-set scenarios. In the future, we aim to improve the zero-shot generalization ability of our approach.}

% \section*{Acknowledgements}
% This work is supported by National Natural Science Foundation of China (62302093, 62276134, 52441503), Jiangsu Province Natural Science Fund (BK20230833), Double First-Class Construction Foundation of China under Grant 23GH020227, and Big Data Computing Center of Southeast University.

%%%%%%%%%%%%%%%%%%%%%%%%%%%%%%%%%%%%%%%%%%%%%%%%%%%%%%%%%%%%%

\ifCLASSOPTIONcaptionsoff
  \newpage
\fi

{
	\bibliographystyle{IEEEtran}
	\bibliography{IEEEabrv,IEEEtran.bib}
}

\end{document}